\documentclass[journal]{IEEEtran}
\usepackage{amsmath,amsfonts}
\usepackage{algorithmic}
\usepackage{algorithm}
\usepackage{array}
\usepackage[caption=false,font=normalsize,labelfont=sf,textfont=sf]{subfig}
\usepackage{textcomp}
\usepackage{stfloats}
\usepackage{url}
\usepackage{verbatim}
\usepackage{graphicx}
\usepackage{cite}

\usepackage{xcolor}
\usepackage{comment}
\usepackage{svg}
\usepackage{amssymb}
\usepackage[caption=false]{subfig}
\usepackage{float}
\usepackage{lipsum}
\usepackage{hyperref}
\newcolumntype{C}[1]{>{\centering\arraybackslash}p{#1}}

\hypersetup{colorlinks=true, unicode=true, linkcolor=[rgb]{0.10,0.05,0.67}, citecolor=[rgb]{0.10,0.05,0.67}, filecolor=[rgb]{0.10,0.05,0.67}, urlcolor=[rgb]{0.10,0.05,0.67}}
\graphicspath{{Images/}}

\begin{document}

\title{Compliant Non‐Prehensile Pushing Manipulation}

\author{Francesco Cufino$^{1}$,~\IEEEmembership{}
        Mario Selvaggio$^{1}$,~\IEEEmembership{}
        Fabio Amadio$^{2}$,~\IEEEmembership{}
        and 
        Fabio Ruggiero$^{1}$~\IEEEmembership{}
\thanks{$^{1}$ The authors are with the PRISMA Lab, department of Electrical Engineering and Information Technology of the University of Naples Federico II, Via Claudio, 21, Naples, 80125, Italy. Corresponding author e-mail:
 francesco.cufino@unina.it 

$^{2}$ The author is with Inria, CNRS, Université de Lorraine, Nancy, 54000, France (work carried out while at the ABB Corporate Research Center, Västerås, 72226, Sweden). %
}

\thanks{The research leading to these results has been supported by 
``Dottorati Transizione Digitale" - CUP: E66E23001030002, funded by MUR under the program PNRR - DM 118/2023 Mis.: I.3.4, funded by the European Union Next-Generation EU; 
the ROMOTIVE project, funded by MUR under program FIS3 - DD 1802/2024, grant number FIS-2024-05915.
}

 \thanks{Manuscript received ...; revised ....}

}

\maketitle

\begin{abstract}
In this paper, we address the challenge of performing non-prehensile pushing operations with a compliant robotic manipulation system. 
To ensure safe operations in human-populated environments, robots must comply with external physical interactions and exhibit a passive behavior. 
%
To achieve this, we extend a state-of-the-art pushing model to integrate it with impedance-controlled robots. 
We develop a model predictive control framework built upon this model that enables performing compliant pushing through optimal modulation of the robot's position/velocity set-point, jointly realizing the required pushing force and contact point adaptation to obtain a desired object motion.
However, external interactions may induce tracking errors causing a consequent potentially indefinite increase of the pushing force.
To prevent this, we integrate an energy tank passivity filter that further modulates the robot velocity set-point to guarantee passivity and avoid uncontrolled energy buildup.
The proposed method has been rigorously tested in simulation and validated through experiments on two different robotic systems, demonstrating passive compliance during human–robot interactions, and assessing trajectory tracking performance and robustness to variations in the object’s physical parameters.
\end{abstract}


\section{Introduction}

\IEEEPARstart{S}{ervice} robots are increasingly deployed in human-populated environments, where they are required to both manipulate objects and ensure safe operation under intentional or accidental physical contacts with nearby people~\cite{Lee}. 
Across a wide range of contexts, spanning from professional to domestic and personal settings, robots are expected to operate in close proximity to humans, often sharing workspaces and physically interacting with them during the execution of manipulative tasks.
\begin{figure}[t]
\centering
\includegraphics[width=\linewidth]{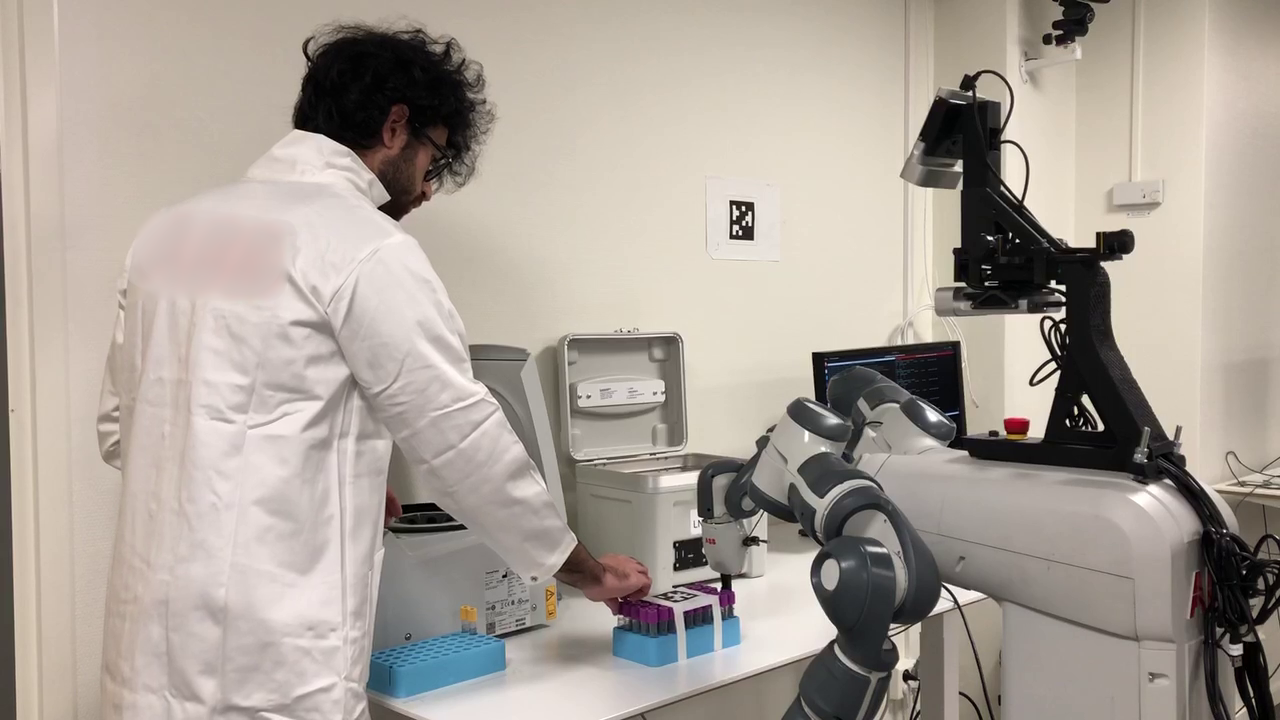}
\caption{Picture showing the envisioned scenario: a robot pushes a rack of test tubes in a hospital environment while a human operator safely interacts with it thanks to our proposed compliant non-prehensile pushing control framework.}
\label{fig:figure1}
\end{figure}
Representative scenarios include 
collaborative industrial workstations, where robots and human operators jointly handle parts, tools, or mobile fixtures; 
assistive settings, where robots support individuals with reduced mobility by repositioning objects or mobility aids; 
and logistics applications, where robots operate in close collaboration with human workers. 
In all these settings, robots must combine \emph{human-like dexterous manipulation} capabilities with \emph{inherently compliant} and \emph{passive} behaviors to enable dependable physical human–robot interaction. 
This work focuses on \emph{non-prehensile pushing} as the manipulation primitive of interest, and develops a control framework that enables service robots to perform pushing tasks in human-centered environments while ensuring safe and compliant behavior during external physical interactions.
As an illustrative example, we focus on a hospital logistics task where a robot delivers a rack of test tubes by pushing it on a horizontal table while interacting with healthcare personnel who may also oppose its motion during operation (see Fig.~\ref{fig:figure1}).
The hospital represents a suitable example of an unstructured, human‐populated environment where other complex manipulation tasks, currently performed by skilled human operators, consist of transporting and delivering medical supplies or serving meals to patients in bed~\cite{Holland}. 
\subsection{Motivation and objective}
In human-populated environments, robots performing manipulation tasks must also guarantee safe physical interaction.
Among the manipulation strategies relevant to these settings, non-prehensile pushing is particularly important, as it enables versatile object repositioning when grasping is impractical (e.g., in case of large or fragile items or constrained workspaces)~\cite{Ruggiero, SelvaggioTCST2023}.
Several methods for non-prehensile pushing manipulation have been proposed in the literature, including model-based approaches that explicitly capture contact mechanics and constraints.
Among these, the recently proposed model predictive control (MPC) framework with complementarity constraints in~\cite{MouraICRA2022} enables pushing via real-time switching between sticking and sliding contact modes. 
However, existing pushing approaches, including~\cite{MouraICRA2022}, are typically designed for position-controlled robots and do not account for physical interaction occurring in shared human–robot environments. 
Extending these methods to impedance-controlled robots is a natural step towards safe coexistence, yet it raises new challenges: the compliance itself may degrade pushing accuracy, and without proper safeguards, external interactions might cause uncontrolled energy buildup in the system, potentially leading to unbounded interaction forces.
Motivated by these considerations, our objective is to develop a control framework for non-prehensile pushing that is suitable for impedance-controlled robots and that jointly enforces passive and compliant behavior with respect to external interactions, while still accomplishing the manipulation task with high precision.

\subsection{Methodology towards the objective}

In the envisioned scenario, the following two requirements must be satisfied \emph{simultaneously} and in a principled way: $(i)$ achieving the manipulation objective with high precision, 
and $(ii)$ guaranteeing a passive and compliant response of the robot to external interactions (e.g., with humans). 
To address the requirement~$(i)$, we propose an extension of the non-prehensile pushing framework based on MPC with complementarity constraints presented in~\cite{MouraICRA2022}, enabling the use of an impedance-controlled robot to perform pushing manipulation tasks.
Our extension generates an optimal position/velocity set-point for the impedance-controlled robot, jointly achieving the required pushing force and contact point evolution to produce the intended object motion, while preserving compliant interaction behavior.
However, compliance alone does not guarantee safe interaction. 
This is precisely why requirement $(ii)$ is essential. 
A key challenge arises when external disturbances, such as human interaction, prevent accurate tracking of the desired object motion. 
In such cases, naive error compensation performed by the MPC framework may produce control actions for the robot which result in potentially unbounded interaction forces and external energy exchange. 
To address this problem, we adopt a passivity-based perspective to ensure that the system does not generate more energy than it receives~\cite{Ott}.
In particular, we incorporate a virtual energy tank in our control framework~\cite{Duindam, Secchi2, Secchi}, that acts as a passivity filter: it monitors the system energy and modifies the optimal set-point when passivity is at risk of being violated. 
This ensures a safe interaction behavior by bounding the energy exchange.

\subsection{Contributions}
In summary, the contributions of this paper are:
\begin{itemize} 
\item A compliant pushing MPC formulation that enables an impedance-controlled robot to achieve the desired object motion by jointly optimizing the pushing force and contact point evolution.
\item An energy tank passivity filter that guarantees bounded energy exchange during non-prehensile pushing in the presence of external interactions.
\item Open-source release of the developed control framework at the following link: \url{https://github.com/prisma-lab/compliant-nonprehensile-pushing-manip}.
\end{itemize}
The devised framework is tested and evaluated in a physics-engine simulator and on two real robotic systems, quantifying its performance and robustness against variability of object's physical parameters and trajectories.

\section{Related works}

In non‐prehensile manipulation, the object is not firmly grasped by the robot and inertial or external forces must be conveniently exploited during manipulation~\cite{Mason3}. 
This allows performing human‐like manipulation primitives such as throwing~\cite{Satici}, catching~\cite{Schill}, batting~\cite{Liu}, pushing~\cite{Bertoncelli}, rolling~\cite{Serra2}, etc. 
Planning and executing them in sequence allows achieving complex manipulation behaviors~\cite{Lynch4}. 
We focus our literature review on pushing, a simple yet highly effective non-prehensile primitive central to our envisioned case study.
%
\subsection{Quasi-static pushing models}
Several papers investigated the theoretical aspects of non‐prehensile pushing. 
The first derivation of a physics‐based model was shown in~\cite{Mason}, where a mapping from the center of pressure to the direction of the object’s angular velocity was established. 
It was soon recognized that the pushed object could move in an unpredictable way due to unknown and variable pressure distribution between an object and its support surface~\cite{Yu2016MoreTA}.
Stability and controllability analyses were conducted to find pushing directions causing the object to remain stably in contact with the manipulator~\cite{Lynch1}.

Quasi‐static analytical models of robotic pushing provide a good and computationally efficient representation useful not only for analysis but also for planning and control.
In quasi‐static conditions, pushing forces can be related to object velocities via a physics‐based criterion known as limit surface, defined as the bounded convex set of friction loads that the surfaces in contact might exhibit~\cite{Goyal}.
The ellipsoidal approximation of the limit surface, introduced in~\cite{Lee_Cutkosky}, yields a differentially flat system~\cite{Zhou2019}, which allows finding invertible models from force to motion~\cite{Lee_Cutkosky} useful for planning~\cite{Mason}, state estimation~\cite{Yu2016MoreTA}, dynamics learning~\cite{Zhou2016}, and feedback control~\cite{Lynch3, Hogan1, Hogan2}. 
\subsection{Switching contact modes}
Although pushing trajectories can be effectively computed by constraining robot‐object interactions to stick, the capability to reason across contact modes leads to more efficient motions and faster corrections~\cite{Zhou2019}.
There are many ongoing efforts to develop local/global motion planning frameworks that can effectively handle the complexity associated with changing frictional contact points~\cite{graesdal2024}. 
These approaches, however, have large computational requirements associated with solving nonlinear and non‐convex optimization programs, unsuitable for online planning and control.
Some recent approaches tend to separate the optimal contact mode sequence, computed offline, from the online search of optimal control inputs computed via MPC used in tandem with integer programming~\cite{Hogan3, Hogan2}. 
Planar pushing allows for flexible contact adaptation by transitioning between two fundamental modes: sticking and sliding. 
To capture the hybrid nature of this interaction, complementarity constraints can be integrated into a MPC formulation and subsequently relaxed, circumventing the need for mixed-integer programming. 
This approach enables real-time mode transitions, facilitating stable pushing operations while simultaneously sliding along the object surface~\cite{MouraICRA2022}.
%

\subsection{Compliant pushing}
To showcase pushing performance, all the previous works use position‐controlled manipulators, which are precise but stiff and prove inadequate when unexpected interactions occour.
Physical constraints imposed by the dynamic contact interaction in manipulation can be handled via impedance control~\cite{HoganACC1984}.
When this is solely used to replicate spring behavior the manipulator exhibits compliance.
Nowadays, the compliant mode can be enabled as a default controller in several commercially-available robotic manipulators.  
This allows specifying the compliance parameters directly in the Cartesian coordinates of the operational space~\cite{KhatibJRA1987}. 
So far, impedance control has been used in several human-robot interaction scenarios, e.g., for implementing walk-through robot programming~\cite{FerragutiFAIM2017}, displaying virtual guides to the user of a telerobotic system~\cite{SelvaggioRAL2018}, and in a unified combination with force control to perform polishing tasks~\cite{HaddadinIJRR2024}. 
More recently, MPC started to be leveraged to synthesize impedance control, handling external interactions while respecting state/input constraints~\cite{BednarczykICRA2020, GoldTRO2023}.
By augmenting the state space of the manipulator system it is also possible to feed back effort measurements useful for explicit force regulation~\cite{KleffIROS2022, SelvaggioTCST2023}.
This is useful to realize a compliant behavior while performing a non‐prehensile manipulation task, enabling a safe response in case of unknown external interactions with humans/obstacles.
However, specifying motion set‐points without jeopardizing the performance of the manipulation task requires particular attention.
An effort to achieve pushing using an impedance-controlled manipulator is presented in~\cite{Bhat}, aiming to prevent unsafe behavior in the presence of uncertainties, which may arise from inaccurate knowledge of the environment (e.g., the surface supporting the object).
Combining this with a dedicated gripper design, a human‐inspired pushing strategy was developed, allowing the contact to be maintained more easily. 
However, such a control strategy has not been conceived to track a planned object trajectory thus severely limiting the manipulation performance.
%
%
%

\subsection{Passivity-based control}
In general, a dynamic interaction among the robot, the environment and/or the human can inject energy into the controlled robotic system. 
This has been observed when realizing a time‐varying impedance~\cite{FerragutiIJRR2019}, or in the presence of switching contacts during manipulation~\cite{ShahriariHUMANOIDS2017, ShahriariRAL2020}.
When an external interaction hinders the robot motion during trajectory tracking, the robot controller reacts to reduce the task error generating internal energy in the system which can potentially grow unbounded~\cite{HaddadinIJRR2024}. 
To preserve a passive behavior, the introduction of energy flow monitoring elements in the controller, also known as energy tanks, was initially proposed in~\cite{Duindam}. 
Energy tanks allow using the dissipated energy to perform potentially non‐passive interactions while still preventing unbounded internal energy generation~\cite{Secchi2, Dietrich}.
The energy tanks passivity‐based control approach has been successfully used so far to realize variable impedance/admittance control in surgical robotics~\cite{SelvaggioRAL2018}, or  control of mobile robots interacting with a dynamic environment~\cite{Brunner}. 

\begin{figure*}
    \centering
    \includegraphics[width=0.75\textwidth]{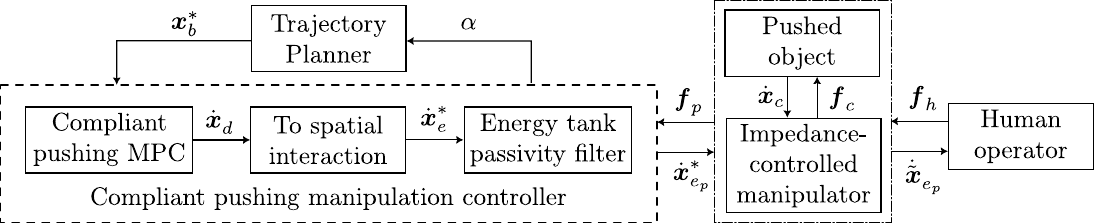} 
    \caption{Block scheme of the considered control framework. 
    Our proposed compliant pushing manipulation controller (dashed box) is composed by a planar compliant pushing MPC and an energy tank passivity filter connected by a planar to spatial set-point transformation. 
    The controller interacts with the impedance-controlled manipulator/object system (dot-dashed box) via a power port and simultaneously acts on the trajectory planner to passively handle external interactions with a human operator.
    All the blocks and symbols are explained throughout the text of Section~\ref{Sec:methods}.
    }
    \label{control_scheme}
\end{figure*}

\section{Control framework}\label{Sec:methods}
This section presents the methodology adopted to realize the sought  compliant and passive pushing behavior. 
Section~\ref{pusher‐slider_model} reviews the pusher–slider model—describing the planar interaction between a point pusher and a sliding object—from~\cite{Hogan1, Hogan2}, together with the complementarity-constraints MPC formulation introduced in~\cite{MouraICRA2022}. 
Building on these foundations, the following subsections describe the control framework that enables an impedance-controlled robot to perform pushing tasks while remaining compliant and passive with respect to external interactions. 
The architecture of the proposed framework is shown in Fig.~\ref{control_scheme}.
Our compliant pushing manipulation controller (within dashed rectangle) comprises three main components. 
First, a \emph{Compliant pushing MPC} block that computes a planar velocity set-point for the robot, relying on a model that extends the one in~\cite{MouraICRA2022} to impedance-controlled robots (Sections~\ref{compliant_pushing_model},~\ref{Problem_formulation}). 
%
%
Second, a \emph{To Spatial Interaction} block lifts the planar velocity set-point to a spatial twist set-point\footnote{Throughout this paper, a \emph{twist} provides a description of the instantaneous velocity of a rigid body in terms of its linear and angular components. Similarly, a \emph{wrench} is a generalized force acting on a rigid body and consists of a linear (force) and an angular component (torque) acting at a point. Depending on the context, twists and wrenches are 3D or 6D vectors~\cite{Murray94}.} (Section~\ref{planar_to_spatial}) suitable for Cartesian impedance-controlled robot manipulators (Section~\ref{impedance_control}). 
Third, before being supplied to the impedance controller, the twist set-point is processed by an \emph{energy tank passivity filter} (Section~\ref{passivity}), to prevent uncontrolled internal energy generation when external interactions occur. 
In addition to these fundamental components, a trajectory planner is assumed to provide the reference trajectory for the object to be pushed. 

As shown on the right part of Fig.~\ref{control_scheme}, the framework explicitly accounts for physical interactions between the controlled robot, the object, and a human operator, all represented as energy-exchanging ports in the block diagram. 
All the notation and variables labeled  are introduced in the corresponding subsections.

\subsection{Planar pushing model} 
\label{pusher‐slider_model}
\begin{figure}[t]
\centering
\includegraphics[width=0.5\linewidth]{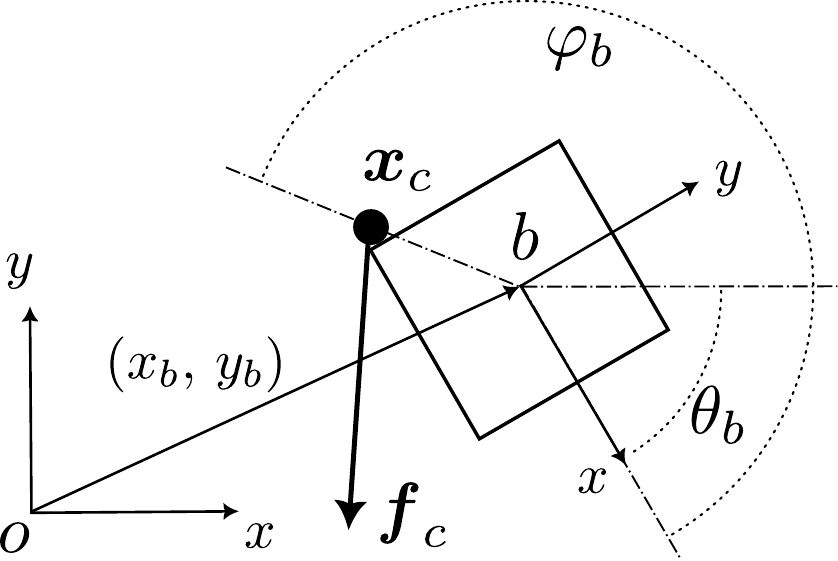}
\caption{Planar pushing model of a rectangular object pushed via a contact force. The symbols in the figure are defined throughout the text of Section~\ref{pusher‐slider_model}. 
}
\label{fig:planar_pushing_model}
\end{figure}

Let us consider a planar pusher-slider system composed of a slider—a rectangular object, assumed to be a rigid body moving on a planar surface—and a pusher—modeled as a point that applies a contact force on the object to control its motion, as shown in Fig.~\ref{fig:planar_pushing_model}. 
In this work, the pusher corresponds to the contact point between a robot manipulator and the object, denoted with $\boldsymbol{x}_c \in \mathbb{R}^2$.
Let us define an inertial, world planar frame, with origin in \(o\), and a body planar frame, with origin in \(b\), attached to the object at its center of mass, as shown in Fig.~\ref{fig:planar_pushing_model}. 
In this case, the system coordinates can be defined as
$
    \boldsymbol{x}_b= \left[x_b \quad y_b \quad \theta_b \quad \varphi_b  \right]^T,
$
where \(x_b, \, y_b \in \mathbb{R}\) and \(\theta_b \in \mathbb{S}^1\) are, respectively, the two position coordinates of the object body and its orientation with respect to the world frame. The coordinate \(\varphi_b \in \mathbb{S}^1\) is defined as the angle between the x-axis of the body frame and the line connecting the body frame origin \(b\) to the contact point $\boldsymbol{x}_c$. 
The model of the considered pusher‐slider system is useful to make the object position and orientation $x_b, y_b, \theta_b$ track a reference time‐varying trajectory expressed in terms of desired quantities $x_b^*(t), \, y_b^*(t) \in \mathbb{R}$ and $\theta_b^*(t) \in \mathbb{S}^1$.
Let us define the object twist in body frame 
$\boldsymbol{v}_b^b \in \mathbb{R}^3$, as
$
    \boldsymbol{v}_b^b= \left[\dot{x}_b^b \quad \dot{y}_b^b \quad \dot{\theta}_b^b\right]^T,
$
where $\dot{x}_b^b, \, \dot{y}_b^b, \, \dot{\theta}_b^b \in \mathbb{R}$ are the two linear body velocity components and the angular one, respectively, and the body wrench $\boldsymbol{f}_b^b \in \mathbb{R}^3$ applied at the body frame origin $b$ due to the pushing force as
$
    \boldsymbol{f}_b^b= \left[f_{b_x}^b \quad f_{b_y}^b \quad \tau_{b_z}^b\right]^T,
$
where $f_{b_x}^b,\, f_{b_y}^b,\, \tau_{b_z}^b \in \mathbb{R}$ are the two linear force and torque components expressed in body frame, respectively.
As for previous works~\cite{MouraICRA2022, Lynch3, Hogan1, Hogan2}, the object planar model in body frame, under quasi-static motion assumption, i.e., neglecting inertial effects, is described by the following equation 
\begin{equation}
\label{quasi_static}
    \boldsymbol{f}_b^b = - \boldsymbol{f}_s^b, 
\end{equation}
where \(\boldsymbol{f}_s^b \in \mathbb{R}^3\) is the wrench due to the sliding friction. 
%
To describe the object motion due to the pushing force, it is necessary to find a mapping between \(\boldsymbol{f}_b^b\) and \(\boldsymbol{v}_b^b\). 
In quasi‐static conditions, these quantities can be related via a physics‐based criterion known as limit surface, defined as the bounded convex set of friction loads that the object surfaces in contact with the ground might exhibit~\cite{Goyal}.
We use the ellipsoidal approximation of the limit surface introduced in~\cite{Lee_Cutkosky}, defined as
$
    H(\boldsymbol{f}_b^b) = \frac{1}{2}\boldsymbol{f}_b^{b^T}\boldsymbol{L} \boldsymbol{f}_b^b,
$
with \(\boldsymbol{L} \in \mathbb{R}^{3 \times 3}\) being a positive-definite matrix~\cite{Zhou2016}. 
The semi‐principal axes of the ellipsoidal limit surface are given by $f_{max}$ and $\tau_{max}$ computed as 
\begin{align*}
    &f_{max} = \mu_g m g, \quad 
    &\tau_{max} = \dfrac{\mu_g m g}{A} \int_{\mathcal{A}} ||\boldsymbol{r}^{dm}|| dA.
\end{align*}
In these expressions, $\mu_g>0$ is the friction coefficient between the object and the ground, $m>0$ is the object mass, $g>0$ is the gravitational acceleration, $A > 0$ is the surface area of the object exposed to friction, $\boldsymbol{r}^{dm} \in \mathbb{R}^2$ is the position of an infinitesimal mass $dm$ relative to the body frame origin \(b\).
By the principle of maximal dissipation, the object reacts to an applied wrench by moving in the direction that maximizes the dissipated power at the frictional interface~\cite{Hogan2}. 
This implies that the resulting object twist is directed along the perpendicular direction to the ellipsoidal limit surface, yielding the following sought mapping between \(\boldsymbol{f}_b^b\) and \(\boldsymbol{v}_b^b\)
\begin{equation} \label{lim_sur}
    \boldsymbol{v}_b^b = \nabla H(\boldsymbol{f}_b^b) = \boldsymbol{L} \boldsymbol{f}_b^b.
\end{equation}
To derive the complete pusher–slider model, it is now necessary to express the wrench $\boldsymbol{f}_b^b$ at the body frame origin as a function of the contact force applied by the pusher on the object.
Let us define the planar force at the contact point expressed in the body frame
\begin{equation} \label{contact_forces}
    \boldsymbol{f}_c^b=[f_{c_x}^b \quad f_{c_y}^b]^T,
\end{equation}
where $f_{c_x}^b >0, \, f_{c_y}^b \in \mathbb{R}$ are the two linear force components that are normal and tangential to the object edge, respectively. 
The contact force $\boldsymbol{f}_c^b$ can be transformed in the equivalent wrench exerted at the body frame origin $\boldsymbol{f}_b^b$ through the transpose of the contact Jacobian matrix 
$\boldsymbol{J}_c \in \mathbb{R}^{2 \times 3}$, i.e.,
\begin{equation} \label{dyn_push_slid_1}
\boldsymbol{f}_b^b = \boldsymbol{J}_c^T \boldsymbol{f}_c^b, \quad \text{with} \quad \boldsymbol{J}_c = \begin{bmatrix}
    1 & 0 & -y_c^b\\
    0 & 1 & x_c^b
    \end{bmatrix},
\end{equation}
in which \(x_c^b,\, y_c^b \in \mathbb{R}\) are the contact point coordinates expressed in body frame.
Plugging~\eqref{dyn_push_slid_1} into~\eqref{lim_sur}, the contact force $\boldsymbol{f}_c^b$ can be directly related to the resulting object twist $\boldsymbol{v}_b^b$.
We denote with $\boldsymbol{v}_b = \boldsymbol{A}_b \boldsymbol{v}_b^b$ the \emph{hybrid} representation of the body twist $\boldsymbol{v}_b^b$ in the world frame~\cite{BRUYNINCKX1996135}, with 
\begin{equation}
\label{dyn_push_slid_3}
    \boldsymbol{A}_b = \begin{bmatrix}
    \boldsymbol{R}_b(\theta_b) & \boldsymbol{0}_2\\ \boldsymbol{0}_2^T & 1
\end{bmatrix},
\end{equation}
where $\boldsymbol{0}_2 \in \mathbb{R}^2$ is the two-dimensional zeros column vector, and $\boldsymbol{R}_b(\theta_b) \in SO(2)$ is the rotation matrix of the body planar frame with respect to the world.
By combining~\eqref{lim_sur},~\eqref{dyn_push_slid_1} and~\eqref{dyn_push_slid_3}, the time derivative of the system coordinates can be described as follows
\begin{equation}
\label{pusher‐slider_dyn}
    \dot{\boldsymbol{x}}_b = 
    \begin{bmatrix}
        \boldsymbol{A}_b \boldsymbol{L} \boldsymbol{J}_c^T \boldsymbol{f}_c^b\\
        \dot{\varphi}_b
    \end{bmatrix}.
\end{equation}
In~\eqref{pusher‐slider_dyn}, the contact point coordinates \(x_c^b,\, y_c^b \in \mathbb{R}\), used in the contact Jacobian~\eqref{dyn_push_slid_1}, 
are parametrized by the current $\varphi_b$ according to the object geometry.
%
An expression for these coordinates can be derived by assuming a rectangular object, whose body frame axes are aligned with its sides, with its origin $b$ coincident with the geometric center of the object, as shown in Fig.~\ref{fig:planar_pushing_model}. 
We further assume that the pusher only contacts the side parallel to the $y$-axis in the negative $x$ half-plane. Denoting by $l>0$ the length of the sides parallel to the $x$-axis, the contact point coordinates are computed as
%
%
\begin{equation}\label{x_c_def}
    \boldsymbol{x}_c^b(\varphi_b) = 
    \begin{bmatrix}x_c^b \quad y_c^b \end{bmatrix}^T =
    \begin{bmatrix} - \dfrac{l}{2} \quad - \dfrac{l}{2} \tan \varphi_b\end{bmatrix}^T.
\end{equation}
If the object geometric center does not coincide with \(b\), an offset between the two should be accounted for in the computation of $\boldsymbol{x}_c^b(\varphi_b)$.
For objects having more general shapes, i.e., non-rectangular or non-convex, particular attention must be devoted to the choice of system coordinates as contact location may not correspond to a unique or well-defined point. 
In this case, one can adopt the choice reported in~\cite{FedericoRAL2025}, where arc-length is used instead of \(\varphi_b\) to parametrize the contact point position along the object contour.

The model in~\eqref{pusher‐slider_dyn} describes the continuous quasi-static behavior of the pusher–slider system. However, the system also exhibits a hybrid nature, as different contact modes can occur between the pusher and the slider during manipulation. 
Following the formulation in~\cite{MouraICRA2022, Hogan3}, assuming the pusher can only exert a uni-directional force on the object, i.e., $f_{c_x}^b\geq0$, three contact modes must be considered, i.e., 
\begin{equation}
\label{contact_modes}
c_1:
\begin{cases}
    \dot{\varphi}_b = 0,\\
    |f_{c_y}^b| \leq \mu f_{c_x}^b,
\end{cases}
\hspace{-0.3cm}
c_2:
\begin{cases}
    \dot{\varphi}_b > 0,\\
    f_{c_y}^b = \mu f_{c_x}^b,
\end{cases}
\hspace{-0.3cm}
c_3:
\begin{cases}
    \dot{\varphi}_b < 0,\\
    f_{c_y}^b = -\mu f_{c_x}^b.
\end{cases}
\end{equation}
The first contact mode $c_1$ is a sticking contact, in which $\varphi_b$ is constant and the applied force belongs to the friction cone characterized by the friction coefficient \(\mu >0\) between pusher and object.
The second contact mode $c_2$ is sliding counterclockwise, in which the applied
force belongs to one edge of the friction cone while \(\dot{\varphi}_b\) is positive. 
The third contact mode $c_3$ is sliding clockwise, in which the applied force belongs to the other edge of the friction cone and \(\dot{\varphi}_b\) is negative. 

%
In the state-of-the-art framework~\cite{MouraICRA2022}, the contact modes are incorporated as constraints in the MPC formulation, which optimizes $\boldsymbol{f}_c^b$ and $\dot{\varphi}_b$ to track a desired object trajectory. 
From these, the reference contact point is derived using~\eqref{x_c_def}. 
With position-controlled robots, pushing can be achieved by directly tracking this reference contact point with high external disturbance rejection.
However, position control makes the robot arbitrarily stiff to external interactions, limiting its deployment in human-populated environments.
In the next subsection, we overcome this limitation by introducing our compliant non-prehensile pushing model.

\subsection{Compliant pushing model}
\label{compliant_pushing_model}
\begin{figure}[t]
\centering
\includegraphics[width=0.5\linewidth]{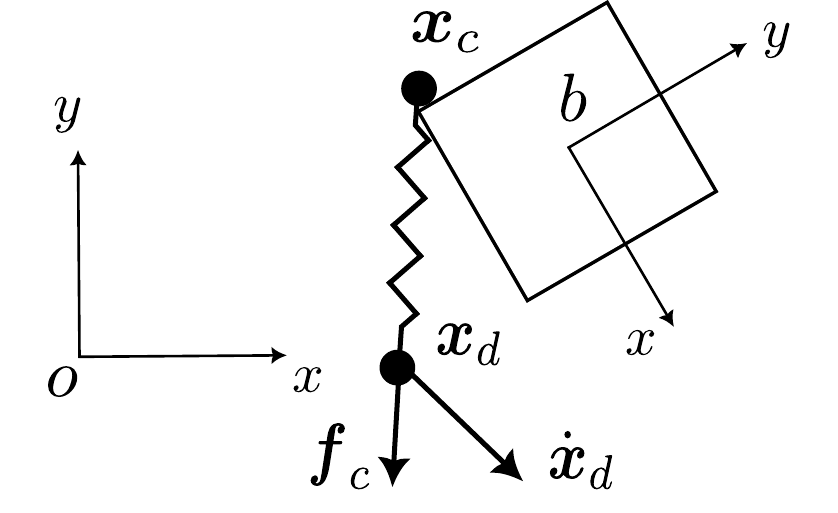}
\caption{Compliant pushing model of an object pushed by a spring-like force $\boldsymbol{f}_c$, applied at the contact point $\boldsymbol{x}_c$. The spring end-point $\boldsymbol{x}_d$ has velocity $\dot{\boldsymbol{x}}_d$, which mediates between different control objectives.}
\label{fig:compliant_pushing_model}
\end{figure}
%
%
%

As discussed in the previous section, in the position-controlled framework~\cite{MouraICRA2022} the MPC produces a predicted contact point, which the robot tracks directly with high stiffness. With an impedance-controlled robot, this approach is no longer viable: a displacement between the set-point and the actual contact point is needed to generate the pushing force required by the task. This calls for a different model that explicitly accounts for the compliant interaction.
To this end, we model the interaction between the robot and the object as a virtual spring (see Fig.~\ref{fig:compliant_pushing_model}). This model captures the quasi-static force-displacement relationship used by the MPC for prediction: under quasi-static conditions, the stiffness term dominates, while inertial and damping contributions are negligible due to the low accelerations and velocities involved. The damping term, although present in the physical impedance controller for stability purposes (Section~\ref{impedance_control}), does not significantly affect this relationship and is therefore omitted in the pushing model. A pushing force is generated by the displacement between the set-point $\boldsymbol{x}_d \in \mathbb{R}^2$, the free, controlled spring end, and the contact point $\boldsymbol{x}_c \in \mathbb{R}^2$, which is constrained by the object edge. At the same time, to precisely steer the object along a given trajectory, a desired sliding velocity $\dot{\boldsymbol{x}}_c$ of the contact point along the object edge must also be produced by the motion of the controlled spring end.
%
%


The model we are deriving here is finalized to formulate an MPC for the compliant case (Sec.~\ref{Problem_formulation}).
Likewise the position-controlled case, the MPC has to plan the evolution of the contact force and the sliding motion along the object edge. However, in the compliant case, the resulting contact point prediction can no longer be directly tracked, as the interaction is mediated by the spring dynamics. The only directly controllable quantity is the spring endpoint velocity $\dot{\boldsymbol{x}}_d \in \mathbb{R}^2$, which must simultaneously encode both the desired force variation and the desired sliding velocity. This introduces a coupling that is absent in the position-controlled case: the realization of force dynamics and sliding velocity are inherently coupled through $\dot{\boldsymbol{x}}_d$. To make this coupling explicit, we derive the relationship between the spring endpoint velocity and the optimized quantities.


The adopted spring model assumes the force to be proportional to the displacement between the spring position set‐point and the contact point expressed in the body frame, i.e.,
\begin{equation} \label{impedance_body_model}
    \boldsymbol{f}_c^b =\boldsymbol{K}(\boldsymbol{x}_d^b - \boldsymbol{x}_c^b(\varphi_b)),
\end{equation}
with $\boldsymbol{K} \in \mathbb{R}^{2 \times 2}$ a diagonal positive-definite stiffness matrix. 
%
Computing the time derivative of~\eqref{impedance_body_model} and isolating the set‐point velocity term leads to
\begin{equation} \label{des_vel_model}
    \dot{\boldsymbol{x}}_d^b = \underbrace{\boldsymbol{K}^{-1} \dot{\boldsymbol{f}}_c^b}_{\text{Force dynamics contribution}}   + \underbrace{\dot{\boldsymbol{x}}_c^b(\varphi_b, \dot{\varphi}_b)}_{\text{Sliding velocity contribution}},
\end{equation}
where $\dot{\boldsymbol{x}}_c^b(\varphi_b, \dot{\varphi}_b)$ can be computed via time derivative of \eqref{x_c_def} as
\begin{equation*}
    \dot{\boldsymbol{x}}_c^b(\varphi_b, \dot{\varphi}_b) = 
    \begin{bmatrix}\dot{x}_c^b \quad \dot{y}_c^b \end{bmatrix}^T =
    \begin{bmatrix} 0 \quad - \dfrac{l}{2} \dfrac{\dot{\varphi}_b}{\cos^2 \varphi_b}\end{bmatrix}^T.
\end{equation*} 
The quantities $\dot{\boldsymbol{f}}_c^b$ and $\dot{\varphi}_b$ are both adopted as control actions optimized by the proposed compliant pushing MPC introduced in the next section.
%
%
%
%
%
The model obtained in~\eqref{des_vel_model} can be physically interpreted and justified as follows. 
If no variation of the contact point along the edge is required, the second term is $\dot{\boldsymbol{x}}_c^b = 0$, and the velocity set-point is influenced only by the first term related to the force dynamic evolution. 
Instead, when no time variation of the force is required, the first term $\dot{\boldsymbol{f}}_c^b = 0$, and only the sliding contact velocity contributes to velocity set-point. 
If both contributions persist, they are coupled into a set‐point velocity which realizes a combination between the two through the control action.

\subsection{Compliant pushing MPC formulation}
\label{Problem_formulation}

Relying on the model just derived, we can now formulate the MPC problem that realizes the sought compliant non-prehensile pushing framework.
The state of the problem is defined as
\begin{equation}
    \label{state}
    \boldsymbol{x} = \begin{bmatrix}
    \boldsymbol{x}_b^T& 
    \boldsymbol{x}_d^{b^T} &
    \boldsymbol{f}_c^{b^T}
    \end{bmatrix}^T,
\end{equation}
where \(\boldsymbol{x} \in \mathbb{R}^n\), with \(n=8\), is composed by the object state $\boldsymbol{x}_b$, the position set-point $\boldsymbol{x}_d^b$, and the desired force $\boldsymbol{f}_c^b$. 
Notice that the desired force is included in the state because we want to optimize the force time derivative appearing in the compliant pushing model in~\eqref{des_vel_model}. 
Furthermore, this allows realizing and exerting a smoother pushing force, 
and introduces the possibility to realize force feedback as a future development~\cite{KleffIROS2022}.
The dynamic model, resulting from the time derivative of the state in~\eqref{state}, is obtained by considering~\eqref{pusher‐slider_dyn} and~\eqref{des_vel_model} as follows
\begin{equation}
    \label{complete_model_dyn}
    \dot{\boldsymbol{x}}=
    \begin{bmatrix}
    \boldsymbol{A}_b \boldsymbol{L} \boldsymbol{J}_c^T \boldsymbol{f}_c^b\\
    \dot{\varphi}_b\\
    \boldsymbol{K}^{-1} \dot{\boldsymbol{f}}_c^b   + \dot{\boldsymbol{x}}_c^b(\varphi_b, \dot{\varphi}_b)\\
    \dot{\boldsymbol{f}}_c^b
    \end{bmatrix}.
\end{equation}

To formulate a physically grounded MPC based on the model in~\eqref{complete_model_dyn}, the three contact modes introduced in~\eqref{contact_modes} must be included as complementarity constraints~\cite{MouraICRA2022}.
%
To this end, two non-negative auxiliary variables $\dot{\varphi}_+, \, \dot{\varphi}_- \geq 0$ are introduced, such that $\dot{\varphi}_b$ is expressed as their difference, i.e., $\dot{\varphi}_b = \dot{\varphi}_+ - \dot{\varphi}_-$. 
By further defining
$\boldsymbol{\lambda}_v = \left[\lambda_-\quad \lambda_+\right]^T = \left[\mu f_{c_x}^b - f_{c_y}^b\quad \mu f_{c_x}^b + f_{c_y}^b\right]^T$ and $\dot{\boldsymbol{\varphi}}_v = \left[\dot{\varphi}_+ \quad \dot{\varphi}_-\right]^T$, it can be shown that the following conditions simultaneously satisfy all three contact modes in~\eqref{contact_modes}
\begin{equation}
\begin{cases} \label{constraints}
    \dot{\varphi}_+, \dot{\varphi}_-, \lambda_-, \lambda_+  \geq 0,\\
    \boldsymbol{\lambda}_v^T \dot{\boldsymbol{\varphi}}_v  + \epsilon = 0.
\end{cases}
\end{equation}
It is worth noting that in~\eqref{constraints}, $\epsilon \in \mathbb{R}$ is a slack variable that relaxes the strict complementarity condition $\boldsymbol{\lambda}_v^T \dot{\boldsymbol{\varphi}}_v = 0$, ensuring numerical tractability of the optimization problem and allowing the system to continuously switch from one active contact mode to another at any instant in time.

The chosen control input $\boldsymbol{u} \in \mathbb{R}^m$, with $m=5$, naturally follows from this formulation, and consists of $\dot{\varphi}_+$, $\dot{\varphi}_-$, the pushing force derivative $\dot{\boldsymbol{f}}_c^b$, and the slack variable $\epsilon$, i.e.,
\begin{equation}
\label{control_input}
    \boldsymbol{u} =
    \begin{bmatrix}
    \dot{\varphi}_+ &
    \dot{\varphi}_- &
    \dot{\boldsymbol{f}}_c^{b^T} &
    \epsilon
    \end{bmatrix}^T.
\end{equation}
%


%

%
%
The formulated MPC problem has the following form
\begin{align*} 
\min_{\boldsymbol{u}} \quad & J(\boldsymbol{u}, \boldsymbol{x}_k)= m(\boldsymbol{x}(t_k + T)) + \int_{t_k}^{t_k+T} l(\boldsymbol{x}(\tau), \boldsymbol{u}(\tau)) d\tau \\
\textrm{s.t.} 
\quad & \dot{\boldsymbol{x}}(\tau) = \boldsymbol{f}(\boldsymbol{x}(\tau),\boldsymbol{u}(\tau)), \quad \boldsymbol{x}(t_k)=\boldsymbol{x}_k, \\ 
 &\boldsymbol{x}(\tau) \in \mathcal{X}, \quad \boldsymbol{u}(\tau) \in \mathcal{U}, 
\end{align*}
and is solved over the time horizon $[t_k, \, t_k + T]$, with $t_k, \, T \in \mathbb{R}$, starting from the initial condition $\boldsymbol{x}_k \in \mathbb{R}^n$.
Mayer and Lagrange terms are defined as
\begin{align*} 
&m(\boldsymbol{x}(t_k + T)) = \boldsymbol{x}^T(t_k + T)\boldsymbol{W}_x\boldsymbol{x}(t_k + T),\\
&l(\boldsymbol{x}(\tau), \boldsymbol{u}(\tau)) = (\boldsymbol{y}^*(\tau)-\boldsymbol{y}(\tau))^T\boldsymbol{W}_y(\boldsymbol{y}^*(\tau)-\boldsymbol{y}(\tau)),
\end{align*}
with \(\boldsymbol{y}=[\boldsymbol{x}^T \quad \boldsymbol{u}^T]^T\) and \(\boldsymbol{y}^*=[\boldsymbol{x}^{*T} \quad \boldsymbol{u}^{*T}]^T\) the reference state/input trajectory, \(\boldsymbol{W}_x \in \mathbb{R}^{n \times n}\) and \(\boldsymbol{W}_y \in \mathbb{R}^{(n+m) \times (n+m)}\) 
positive-definite diagonal weight matrices.

In our implementation, the reference vector $\boldsymbol{y}^*$ has the first three components different from zero (i.e., reference object position and orientation), and all the other ones equal to zero. 
In this way, the optimizer penalizes the object position and orientation error and the difference from zero for all the other variables. 
The weight corresponding to $\varphi_b$ is always set to zero to let the robot freely slide on the object edge.
The system dynamics \(\dot{\boldsymbol{x}}(\tau) = \boldsymbol{f}(\boldsymbol{x}(\tau),\boldsymbol{u}(\tau))\) is expressed by~\eqref{complete_model_dyn}, and constraints \(\boldsymbol{x}(\tau) \in \mathcal{X}, \, \boldsymbol{u}(\tau) \in \mathcal{U}\) are taken from~\eqref{constraints}, to which lower and upper bounds on the state and control input are added. 
Before solving the MPC problem, \(\boldsymbol{x}_b\) is updated with the measured object state, while the other state variables are updated using the model prediction. 
For the implementation, the MPC is  discretized with a sampling frequency $\nu_s \in \mathbb{R}$ and the prediction time horizon is chosen by considering $N \in \mathbb{N}$ samples. 
All the weights, sampling frequency, and time horizon chosen in our simulations and experiments in Sections~\ref{sec:simulation} and~\ref{sec:experiments} are reported in Table~\ref{tab:param_sim},~\ref{tab:param_yumi}, and~\ref{tab:param_kuka}. 

In summary, the formulated MPC receives as input the reference trajectory and the object state feedback, and computes time derivative of contact force and sliding velocity from which we derive the planar position/velocity set-point \(\boldsymbol{x}_d^b ,\, \dot{\boldsymbol{x}}_d^b \in \mathbb{R}^2\) resulting from~\eqref{des_vel_model} specified with respect to the object frame. 
The velocity set-point is then transformed into the world frame. 
Unless otherwise specified, all the referenced quantities are expressed in the world frame.
\subsection{From planar to spatial interaction}
\label{planar_to_spatial}
This subsection details how the planar MPC set-points and target forces are mapped into 3D Cartesian space to execute compliant pushing via an impedance-controlled manipulator.

\begin{figure}[t]
\centering
\includegraphics[width=\linewidth]{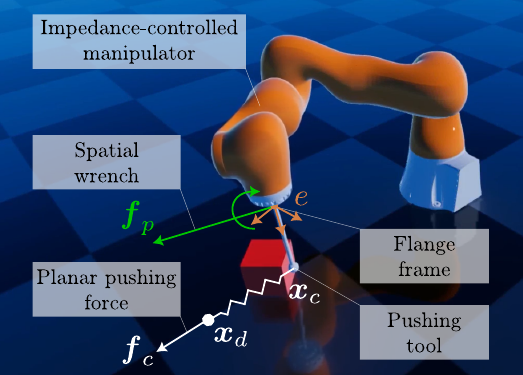}
\caption{Illustration of the proposed compliant non‐prehensile pushing manipulation system with associated 3-D reference frames. The realized planar spring-like pushing force $\boldsymbol{f}_c$ is proportional to the displacement between $\boldsymbol{x_d}$ and $\boldsymbol{x_c}$. $\boldsymbol{f}_p$ is the equivalent spatial wrench in the robot flange frame. 
}
\label{Frames}
\end{figure}

As detailed previously, our compliant pushing MPC strategy modulates the planar position/velocity set-point \(\boldsymbol{x}_d,\, \dot{\boldsymbol{x}}_d\) to indirectly generate an optimal manipulative force $\boldsymbol{f}_c$ at the contact point.
This force is physically realized by the impedance-controlled manipulator and transmitted to the object via a dedicated pushing tool attached to the robot flange—the standardized mounting surface at the end of a robot arm, acting as the crucial interface for attaching end-effectors (i.e., tools like grippers, welders). 
Figure~\ref{Frames} illustrates the relevant system coordinate frames and quantities used throughout the subsequent description.

Through Cartesian impedance control, the robot frame attached to the flange (see Figure~\ref{Frames}) is regulated to match a target impedance dynamics (see Sec.~\ref{impedance_control}). 
Imposing a compliant behavior enables the robot to generate a generalized force in response to commanded Cartesian set-point displacements. 
To produce these displacements, the planar position/velocity set-points—and consequently the desired contact force—must be mapped from the 2D plane to the 3D Cartesian space. 
These variables are expressed as spatial quantities (3D pose and twist for the set-points, and a wrench for the force) and transformed into the flange frame. 

Let us denote with $\boldsymbol{x}_e = (\boldsymbol{p}_e, \boldsymbol{R}_e) \in \mathbb{SE}(3)$ the pose composed of the position and the orientation matrix of the flange frame with origin in $e$ and with $\dot{\boldsymbol{x}}_e \in \mathbb{R}^6$ its twist.
The flange set-point, consisting of the reference pose $\boldsymbol{x}_e^* = (\boldsymbol{p}_e^*, \boldsymbol{R}_e^*) \in \mathbb{SE}(3)$ and its reference twist $\dot{\boldsymbol{x}}_e^* \in \mathbb{R}^6$, is constructed by combining the planar position/velocity set-point in the horizontal plane generated by the MPC, a constant reference height along the vertical $z$ axis, and a fixed orientation that points the pushing tool downward. The spatial pose/twist set-point $\boldsymbol{x}_e^* ,\,\dot{\boldsymbol{x}}_e^*$ is provided to the manipulator impedance controller described in the next section.

%
Dually, the planar contact force $\boldsymbol{f}_c$ applied to the object is mapped to the spatial wrench $\boldsymbol{f}_p \in \mathbb{R}^6$ applied to the flange as
\begin{equation}
\label{pushing_force}
    \boldsymbol{f}_p = \boldsymbol{A}^T \boldsymbol{\Sigma}^T  \boldsymbol{f}_c, \quad 
\end{equation}
where
\begin{equation*}
     \boldsymbol{\Sigma} = \begin{bmatrix}
         1 & 0 & 0 & 0 & 0 & 0 \\
         0 & 1 & 0 & 0 & 0 & 0 
     \end{bmatrix}, \quad \boldsymbol{A}=\begin{bmatrix}
         \boldsymbol{I}_3 & -\boldsymbol{S}(\boldsymbol{p}_{e,c})\\
         \boldsymbol{O}_3 & \boldsymbol{I}_3 \end{bmatrix},
\end{equation*}
and $\boldsymbol{A}^T \in \mathbb{R}^{6 \times 6}$ transforms wrenches between the contact and the flange frames,
with $\boldsymbol{S}(\cdot): \mathbb{R}^3 \rightarrow \mathbb{R}^{3 \times 3} $ the skew-symmetric operator~\cite{Robotics}, $\boldsymbol{I}_\times, \boldsymbol{O}_\times$  the identity and zero matrices of proper dimensions, respectively, and $\boldsymbol{p}_{e,c} \in \mathbb{R}^3$ the distance between the flange and the contact frames expressed in the world frame. 
The transformation introduced in~\eqref{pushing_force} is exploited for the passivity filter introduced later in Sec.~\ref{passivity}.
\label{compliant_pushing}
\subsection{Manipulator impedance control}
\label{impedance_control}
Classical impedance control establishes a desired dynamic relationship between a robot position and the external forces it exerts on its environment.
Instead of strictly controlling position or force independently, the controller makes the robot behave like a mechanical mass-spring-damper system, ensuring compliant and safe physical interactions.
This behavior is generally achieved by choosing the joint control torques $\boldsymbol{\tau}_m \in \mathbb{R}^{n_q}$ (with $n_q \in \mathbb{N}$ denoting the number of joints) as follows:
\begin{equation} \label{control_torques}
\begin{aligned}
    \boldsymbol{\tau}_m &=  \boldsymbol{g}(\boldsymbol{q}) + \boldsymbol{J}(\boldsymbol{q})^T(\boldsymbol{M}_k(\boldsymbol{x}_e) \ddot{\boldsymbol{x}}^*_e +  \boldsymbol{C}_k(\boldsymbol{x}_e, \dot{\boldsymbol{x}}_e)  \dot{\boldsymbol{x}}^*_e +\\
   & \quad + \boldsymbol{K}_d\tilde{\boldsymbol{x}}_e + \boldsymbol{D}_d\dot{\tilde{\boldsymbol{x}}}_e),
\end{aligned}
\end{equation}
With the control commands in~\eqref{control_torques}, we can obtain the following impedance dynamics of the frame attached to the flange expressed in Cartesian coordinates relative to the world frame
\begin{equation} \label{impedance}
    \boldsymbol{f}_e = \boldsymbol{M}_k(\boldsymbol{x}_e)\ddot{\tilde{\boldsymbol{x}}}_e + \left(\boldsymbol{C}_k(\boldsymbol{x}_e, \dot{\boldsymbol{x}}_e) + \boldsymbol{D}_d\right)\dot{\tilde{\boldsymbol{x}}}_e + \boldsymbol{K}_d\tilde{\boldsymbol{x}}_e.
\end{equation}
In the equations above, $\boldsymbol{q} \in \mathbb{R}^{n_q}$ are the joint variables, $\boldsymbol{g}(\boldsymbol{q}) \in \mathbb{R}^{n_q}$ is the gravity, $\boldsymbol{J}(\boldsymbol{q}) \in \mathbb{R}^{6 \times n_q}$ is the Jacobian. The quantity $\tilde{\boldsymbol{x}}_e = (\tilde{\boldsymbol{p}}_e,\, \tilde{\boldsymbol{\phi}}_e) \in \mathbb{R}^6$ is the flange frame position/orientation error, with $\tilde{\boldsymbol{p}}_e = \boldsymbol{p}^*_e - \boldsymbol{p}_e$, and $\tilde{\boldsymbol{\phi}}_e$ the roll, pitch, and yaw Euler angles extracted from the matrix $\boldsymbol{R}_e^*\boldsymbol{R}_e^T$.
The matrices \(\boldsymbol{M}_k(\boldsymbol{x}_e),\,  \boldsymbol{C}_k(\boldsymbol{x}_e, \, \dot{\boldsymbol{x}}_e) \in \mathbb{R}^{6 \times 6} \) are the inertia and Coriolis terms of the robot in Cartesian coordinates, respectively, defined in~\cite{Ott}, and $\boldsymbol{f}_e \in \mathbb{R}^6$ is the external wrench applied at the flange. 
Finally, the positive-definite matrices \(\boldsymbol{D}_d, \, \boldsymbol{K}_d \in \mathbb{R}^{6 \times 6}\) are desired impedance damping and stiffness, respectively.
Without using an external force/torque sensor, from~\eqref{impedance} it is easy to notice that we can freely shape the desired stiffness, $\boldsymbol{K}_d$, we can add a contribution to the damping through $\boldsymbol{D}_d$, while we cannot change the robot's inertia $\boldsymbol{M}_k$ via the control action~\eqref{control_torques}~\cite{Ott}. 
However, in view of the quasi-static conditions,  the last term of~\eqref{impedance} is the major contributor towards the physical realizability of the external wrench $\boldsymbol{f}_e$.
By matching the stiffness of the compliant pushing model introduced in Section~\ref{compliant_pushing_model} along the first two linear positional components, this term will be used to realize the contact force in~\eqref{impedance_body_model}, in view of the transformation~\eqref{pushing_force}.



\subsection{Passivity filter}
\label{passivity}
\begin{figure}[t]
\centering
\includegraphics[width=0.75\linewidth]{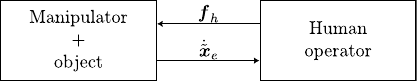}
\caption{Port representation of the compliant pushing system interacting with a human. 
}
\label{Human_manip_port}
\end{figure}

We now analyze the case in which an external interaction occurs during the execution of the non-prehensile pushing task. 
Intuitively, if such an interaction prevents accurate tracking of the object reference trajectory, the MPC would attempt to compensate for the growing tracking error by increasing the pushing force. 
This response is unsafe, and can be interpreted as a potentially unbounded generation of internal energy within the system. 
To study this behavior, we consider a scenario in which a human interacts with the object by applying an external wrench \(\boldsymbol{f}_h \in \mathbb{R}^6\), 
while the robot is pushing the object via the pushing wrench \(\boldsymbol{f}_p\). 
We suppose that the human wrench \(\boldsymbol{f}_h\) is such that it hinders the motion of the object and does not cause a contact loss between the object and the pushing tool, as this could not be counteracted by the manipulator performing a non-prehensile manipulation task~\cite{Ruggiero}.
The considered system can be represented by the block diagram shown in Fig.~\ref{Human_manip_port}, where the block \emph{manipulator + object} describes the controlled manipulator while interacting with the object, receiving as input the human wrench \(\boldsymbol{f}_h\) and producing as output the twist error \(\dot{\tilde{\boldsymbol{x}}}_e\).
We investigate the passivity of the proposed system with respect to the input–output pair given by the interaction wrench and the twist error $(\boldsymbol{f}_h, \dot{\tilde{\boldsymbol{x}}}_e)$.
For the impedance-controlled manipulator, we consider the total external wrench in~\eqref{impedance} decomposed in the sum of the two contributions 
\begin{equation}
    \label{eq:ext_force}
    \boldsymbol{f}_e = \boldsymbol{f}_p + \boldsymbol{f}_h.
\end{equation}
Let us consider a storage function \(V : \mathbb{R}^{12} \rightarrow \mathbb{R}_+\) representing the elastic potential energy plus the kinetic energy of the system 
\begin{equation}\label{eq:storage_syst}
    V = \dfrac{1}{2} \tilde{\boldsymbol{x}}_e^T \boldsymbol{K}_d \tilde{\boldsymbol{x}}_e + \dfrac{1}{2} \dot{\tilde{\boldsymbol{x}}}_e^T \boldsymbol{M}_k(\boldsymbol{x}_e) \dot{\tilde{\boldsymbol{x}}}_e.
\end{equation}
Passivity is enforced by the condition \(\dot{V} \leq \dot{\tilde{\boldsymbol{x}}}_e^T \boldsymbol{f}_h\). 
Computing the time derivative of \(V\), combining it with~\eqref{impedance} and~\eqref{eq:ext_force}, and exploiting the skew-symmetric property of the matrix \(\dot{\boldsymbol{M}}_k(\boldsymbol{x}_e) - 2 \boldsymbol{C}_k(\boldsymbol{x}_e, \dot{\boldsymbol{x}}_e)\)~\cite{Robotics}, we have
%
%
%
$    \dot{V} = \dot{\tilde{\boldsymbol{x}}}_e^T\boldsymbol{f}_h + \dot{\tilde{\boldsymbol{x}}}_e^T\boldsymbol{f}_p -\dot{\tilde{\boldsymbol{x}}}_e^T \boldsymbol{D}_d \dot{\tilde{\boldsymbol{x}}}_e.$
The term representing the energy dissipated by the damping effect is \(-\dot{\tilde{\boldsymbol{x}}}_e^T \boldsymbol{D}_d \dot{\tilde{\boldsymbol{x}}}_e < 0\),  but the term $\dot{\tilde{\boldsymbol{x}}}_e^T\boldsymbol{f}_p$, representing the energy exchanged across the twist error and the pushing wrench port has undefined sign,
which means that energy injection can occur, causing passivity violation.

To solve this problem, we make the MPC output passive through a \emph{passivity filter}.
This is achieved by introducing an energy tank designed to store the energy dissipated by the system, thereby enabling a less conservative control action while preserving the overall passivity of the system.
To this end, an energy storing element \(z \in \mathbb{R}\) is added to the state of the system, whose storage function \(T : \mathbb{R} \rightarrow \mathbb{R}_+\) is
$
    T(z) = 0.5 z^2.
$
An upper bound limitation \(\bar{T}>0\) for the energy stored in the tank is necessary to avoid practical instability issues~\cite{Lee_huang}. 
%
The tank is singular for \(z=0\), hence a lower bound \(T_\epsilon>0\) for the tank energy must also be defined.
The twist set‐point $\dot{\boldsymbol{x}}_e^*$ generated by the MPC is processed through the passivity filter to obtain a passive set-point $\dot{\boldsymbol{x}}_{e_p}^* \in \mathbb{R}^6$. 
This is done to track the potentially non-passive set-point as accurately as possible until a passivity violation occurs.

The energy tank dynamics is defined as
\begin{equation}
\label{tank_dyn}
    \dot{z} = \dfrac{\beta}{z} \dot{\tilde{\boldsymbol{x}}}_{e_p}^T \boldsymbol{D}_d \dot{\tilde{\boldsymbol{x}}}_{e_p} - \dfrac{\gamma}{z} \dot{\tilde{\boldsymbol{x}}}_{e_p}^T  \boldsymbol{f}_p,
\end{equation}
where \(\beta, \gamma \in \{0,1\}\) are parameters used to enforce the upper bound limitation \(\bar{T}\) for the energy stored in the tank. The first term, representing dissipated energy, replenishes the tank, whereas the second term—associated with physical interaction—may either supply or deplete energy.
%
The core idea behind the passivity filter is the following.
When the tank energy falls below the lower threshold $T_{\epsilon}$ and the internal energy is increasing, the twist set‐point no longer tracks the non‐passive set‐point; instead it is set to the actual twist to eliminate the twist error. 
This can be formalized through the following passivity filter 
\begin{equation}
\label{pass_fil_compact}
     \dot{\boldsymbol{x}}_{e_p}^* = \alpha \left(\boldsymbol{\varLambda} \left( \boldsymbol{x}_e^* - \boldsymbol{x}_{e_p}^* \right)  + \dot{\boldsymbol{x}}_e^* \right) + (1 - \alpha) \dot{\boldsymbol{x}}_e,
\end{equation}
with
\begin{equation}
\label{switching_condition}
    \alpha =
    \begin{cases}
    0, & T \leq T_\epsilon \wedge \dot{\tilde{\underline{\boldsymbol{x}}}}_{e_p}^T  \boldsymbol{f}_p > 0,\\
         1 , & \text{otherwise},
    \end{cases}
\end{equation}
and \(\dot{\tilde{\underline{\boldsymbol{x}}}}_{e_p} =\boldsymbol{\varLambda} \left( \boldsymbol{x}_e^* - \boldsymbol{x}_{e_p}^* \right)  + \dot{\boldsymbol{x}}_e^* - \dot{\boldsymbol{x}}_e\).
The term \(\boldsymbol{\varLambda} \in \mathbb{R}^{6 \times 6}\) is a positive semidefinite diagonal gain matrix (in which only the first two elements of the diagonal are non‐null since the other components of the twist set-point are always zero).  
Let us now define the error $\tilde{\boldsymbol{x}}_{e_p} = \boldsymbol{x}_{e_p}^* - \boldsymbol{x}_e$.
Equation~\eqref{pass_fil_compact} implies that, when passivity is violated, \(\dot{\tilde{\boldsymbol{x}}}_{e_p}=0\).
It is worth noting that the quantity $\dot{\tilde{\underline{\boldsymbol{x}}}}_{e_p}$ is used in the switching condition in~\eqref{switching_condition} instead of $\dot{\tilde{\boldsymbol{x}}}_{e_p}$ to make the state stable with $\alpha=0$. 
Indeed, $\alpha=0$ implies $\dot{\tilde{\boldsymbol{x}}}_{e_p}^T  \boldsymbol{f}_p = 0$ while this would imply $\alpha=1$ if $\dot{\tilde{\boldsymbol{x}}}_{e_p}$ was adopted in~\eqref{switching_condition}, instead. 

\begin{figure}[t]
\centering
\includegraphics[width=0.45\textwidth]{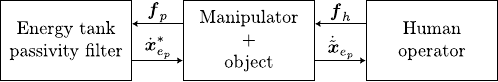}
\caption{Port representation of the compliant pushing system interacting with a human equipped with energy tank passivity filter. 
}
\label{Human_manip_tank_port}
\end{figure}

The system model is now represented by the equations~\eqref{impedance},~\eqref{eq:ext_force},~\eqref{tank_dyn},~\eqref{pass_fil_compact},~\eqref{switching_condition}, where in~\eqref{impedance} the terms $\tilde{\boldsymbol{x}}_e,\, \dot{\tilde{\boldsymbol{x}}}_e,\,\ddot{\tilde{\boldsymbol{x}}}_e$ are replaced by $\tilde{\boldsymbol{x}}_{e_p}, \,\dot{\tilde{\boldsymbol{x}}}_{e_p}, \,
\ddot{\tilde{\boldsymbol{x}}}_{e_p}$, respectively.
The block scheme with port representation of the system equipped with the tank is shown in Fig.~\ref{Human_manip_tank_port}.
It is now possible to introduce a new storage function, obtained through the sum of robotic system~\eqref{eq:storage_syst} and of the tank $T(z)$. 
%
%
The expression of the new storage function \(\mathcal{V} : \mathbb{R}^{13} \rightarrow \mathbb{R}_+\) is
$$
    \mathcal{V}  = V + T = \frac{1}{2}\tilde{\boldsymbol{x}}_{e_p}^T \boldsymbol{K}_d \tilde{\boldsymbol{x}}_{e_p} + \frac{1}{2}\dot{\tilde{\boldsymbol{x}}}_{e_p}^T \boldsymbol{M}_k(\boldsymbol{x}_e) \dot{\tilde{\boldsymbol{x}}}_{e_p} + \frac{1}{2} z^2,
$$
where the error terms are substituted by their passive counterpart.
Computing the time derivative of $\mathcal{V}$ and combining it with~\eqref{tank_dyn} leads to
\begin{equation}
\begin{aligned}
    \dot{\mathcal{V}} &= \dot{\tilde{\boldsymbol{x}}}_{e_p}^T\boldsymbol{f}_h + \dot{\tilde{\boldsymbol{x}}}_{e_p}^T\boldsymbol{f}_p -\dot{\tilde{\boldsymbol{x}}}_{e_p}^T \boldsymbol{D}_d \dot{\tilde{\boldsymbol{x}}}_{e_p} + z \dot{z} =\\
    \label{Vdot3}
    &= (1 - \gamma) \dot{\tilde{\boldsymbol{x}}}_{e_p}^T\boldsymbol{f}_p - (1 - \beta) \dot{\tilde{\boldsymbol{x}}}_{e_p}^T \boldsymbol{D}_d \dot{\tilde{\boldsymbol{x}}}_{e_p} + \dot{\tilde{\boldsymbol{x}}}_{e_p}^T\boldsymbol{f}_h \leq \dot{\tilde{\boldsymbol{x}}}_{e_p}^T\boldsymbol{f}_h.
\end{aligned}
\end{equation}
Observing this storage function, to enforce upper/lower bounds of the tank energy it is sufficient to choose \(\beta\) and \(\gamma\) as 
\begin{align*}
    \beta &= \begin{cases}
        1, & T < \bar{T},\\
        0, & \text{otherwise},
    \end{cases} & 
    \gamma &= \begin{cases}
        \beta, &  \dot{\tilde{\boldsymbol{x}}}_{e_p}^T  \boldsymbol{f}_p < 0,\\
        1, & \text{otherwise}.
    \end{cases}
\end{align*}
The parameter \(\beta\) disables the dissipated energy recovery if an upper bound is reached, in order to avoid practical instability issues~\cite{Lee_huang}.
The parameter \(\gamma\), like  \(\beta\), disables the energy recovery when \(\dot{\tilde{\boldsymbol{x}}}_{e_p}^T  \boldsymbol{f}_p < 0\), corresponding to the condition of increasing tank energy and the upper bound reached.
With these choices, considering all the possible cases, it can be deduced from~\eqref{Vdot3} that the system is passive with respect to the pair (\(\boldsymbol{f}_h,
\dot{\tilde{\boldsymbol{x}}}_{e_p}\)) with storage function $\mathcal{V}$.

It is worth noting that the introduced tank filter requires knowledge of the current pushing wrench $\boldsymbol{f}_p$. 
However, even when measuring the external wrench is possible via a dedicated force/torque sensor mounted at the flange, it is difficult to isolate $\boldsymbol{f}_p$ from $\boldsymbol{f}_e$, as measurements would always contain the total wrench $\boldsymbol{f}_p + \boldsymbol{f}_h$. 
In our implementation, we use as $\boldsymbol{f}_p$ the one derived in~\eqref{pushing_force}, in which $\boldsymbol{f}_c$ is the contact force predicted by the MPC. 
Moreover, to avoid an indefinite growth of the tracking error during interaction, the object reference set-point extracted from the trajectory $\boldsymbol{x}_b^*(t(\alpha))$ is kept constant when $\alpha=0$, making the planner aware of the tank activation.
\section{Simulation}\label{sec:simulation}
%
In this section, we validate the proposed control framework in simulation prior to its deployment on real hardware. 
Simulation allows testing potentially dangerous conditions without inducing any risk to human operators or equipment. 
We analyze the system's response when an external obstacle hinders the object's motion during the pushing task along two comparative cases: one in which the passivity filter is active (\emph{passive case}), and one in which it is removed (\emph{non-passive case}).
By considering these two cases, we aim to demonstrate two key aspects: 
first, that the proposed control scheme maintains high tracking performance in non-prehensile pushing tasks despite the robot operating under impedance control; 
and second, that the passivity filter effectively prevents unsafe force escalation by modulating the twist set-point when external interactions lead to increased tracking error and pushing force.
The simulation setup, including the physics engine, the robot configuration, and the selected parameters, is described in Section~\ref{sec:sim_setup}, while the obtained results are reported and discussed in Section~\ref{sec:results}. 

\subsection{Simulation setup}\label{sec:sim_setup}
The simulation has been implemented in RAISIM\footnote{\url{https://raisim.com/}}, a cross-platform multi-body physics engine for robotics and AI. 
RAISIM provides a realistic contact simulation, particularly beneficial for obtaining reliable results in the considered pushing task. 
The robot we use is the 7‐DOF KUKA LBR iiwa 7, commanded in torque mode and equipped with a pushing tool consisting of a steel stick with a spherical-shaped end.
Its radius, $r = 0.01$~m, is treated as an offset when defining the set-point.
The robot pushes through the tool a steel box whose nominal mass is $m=0.5$~kg and whose dimensions are $l \times l \times l$, with $l = 0.1$~m. 
The box slides on a brass surface. 
The friction coefficients between steel-steel and steel-brass were both set to $\mu_g=0.2$. 
The simulation integration step is $t_i=0.001$~s.
The robot control torques $\boldsymbol{\tau} \in \mathbb{R}^{n_q}$ are set as
$
    \boldsymbol{\tau} = \boldsymbol{\tau}_{task} + \boldsymbol{\tau}_{null},
$
where \(\boldsymbol{\tau}_{task},\,\boldsymbol{\tau}_{null} \in \mathbb{R}^{n_q}\) are, respectively, the task and the null space control torques. 
The task control torques \(\boldsymbol{\tau}_{task}\) are the ones in~\eqref{control_torques} in which the reference acceleration and the Coriolis effect are neglected by virtue of the quasi‐static hypotheses.
The null space control torques are chosen as
\begin{equation*}
    \boldsymbol{\tau}_{null} = \left(\boldsymbol{I}_{n_q} - \boldsymbol{J}^{\dag}_M(\boldsymbol{q}) \boldsymbol{J}(\boldsymbol{q})\right)\left(\boldsymbol{K}_{n} (\boldsymbol{q}_n - \boldsymbol{q}) - \boldsymbol{D}_{n} \dot{\boldsymbol{q}}\right),
\end{equation*}
where \(\boldsymbol{q}_n \in \mathbb{R}^{n_q}\) is the desired nominal configuration chosen equal to the initial one \(\boldsymbol{q}_0 \in \mathbb{R}^{n_q}\), and \(\boldsymbol{K}_{n},\,\boldsymbol{D}_{n} \in \mathbb{R}^{{n_q} \times {n_q}}\) are positive-definite gain matrices. The matrix \(\boldsymbol{J}^{\dag}_M = \boldsymbol{M}^{-1}(\boldsymbol{q})\boldsymbol{J}^T(\boldsymbol{q}) (\boldsymbol{J}(\boldsymbol{q}) \boldsymbol{M}^{-1}(\boldsymbol{q}) \boldsymbol{J}^T(\boldsymbol{q}))^{-1}\) is the inertia-weighted right pseudo-inverse of \(\boldsymbol{J}(\boldsymbol{q})\).
For the MPC implementation, we use the software package \emph{acados}\footnote{\url{https://docs.acados.org/}} which provides fast and embedded solvers for nonlinear optimal control.
%
For both cases, the selected parameters are summarized in Table~\ref{tab:param_sim}. 
The MPC weighting matrices were systematically and iteratively tuned across multiple simulation trials. 
%
Weights on the object tracking error were set as high as possible without jeopardizing the stability of the overall framework, while weights on other state and control variables were maintained at significantly lower magnitudes.
The passivity filter gains were determined empirically, ensuring consistency with the system's energy scale. 
Robot stiffness and damping parameters were selected to reflect realistic conditions in a human-centered environment, where the robot is expected to exhibit compliant behavior during physical interactions with humans.
The state bounds are set to constrain the contact point to the valid region of the object surface, ensuring that it remains inside the object boundary, preventing any violation of the object-edge limits during manipulation.
\begin{table*}[t]
\centering
\caption{Parameters choice of the control framework used for simulations.}
\label{tab:param_sim}
\setlength{\extrarowheight}{2pt}
\begin{tabular}{|p{6cm}|p{6cm}|p{3.5cm}|}
\hline
\centering{\bfseries MPC}&\centering{\bfseries Robot} & \centering\arraybackslash{\bfseries Passivity filter}\\
\hline
$\boldsymbol{x}_0=[0, 0.6, \frac{\pi}{4}, \pi, \frac{-l}{2}, 0, 0, 0]$ 
&
$\boldsymbol{K}_d = \operatorname{diag}\{3\mathrm{e}2,3\mathrm{e}2, 3\mathrm{e}2, 90, 90, 90\}$
&
$T_0 = 1\mathrm{e}{-2}$
\\
$\boldsymbol{w}_x = [1\mathrm{e}6, 1\mathrm{e}6, 1.5\mathrm{e}6, 0, 0, 0, 1\mathrm{e}{-2}, 0.1]$ 
&
$\boldsymbol{D}_d = \operatorname{diag}\{50,50, 50, 15, 15, 15\}$
&
$z_0 = \sqrt{2T_0}$\\
$\boldsymbol{w}_{u} = [1\mathrm{e}{-3}, 1\mathrm{e}{-3}, 1\mathrm{e}{-2}, 1, 10]$ 
&
$\boldsymbol{K}_n = \boldsymbol{I}_{n_q}$
&
$\bar{T} = 1\mathrm{e}{-2}$
\\
$\boldsymbol{W}_x = \operatorname{diag}\{\boldsymbol{w}_x\},\,\boldsymbol{W}_y = \operatorname{diag}\{10\boldsymbol{w}_x, \boldsymbol{w}_{u}\}$
&
$\boldsymbol{D}_n = 0.1 \boldsymbol{I}_{n_q}$
&
$
T_{\epsilon} = 5\mathrm{e}{-4}
$
\\
$\nu_s = 1\mathrm{e}{3}$ Hz, $N = 5$
&
$\boldsymbol{q}_n = [0.85, 1.50, 1.57, -1.65, -1.50, 1.57, 2.50]$
&
$\boldsymbol{\varLambda} = \operatorname{diag}\{50, 50, 0, 0, 0, 0\} $
\\
\hline
\end{tabular}
\end{table*}

\subsection{Results} \label{sec:results}
%
%
\begin{figure}[t]
    \centering
    \captionsetup[subfloat]{labelfont=scriptsize,textfont=scriptsize}
    \subfloat[]{\includegraphics[clip, trim=0cm 2.7cm 0cm 1.1cm,width=0.48\textwidth]{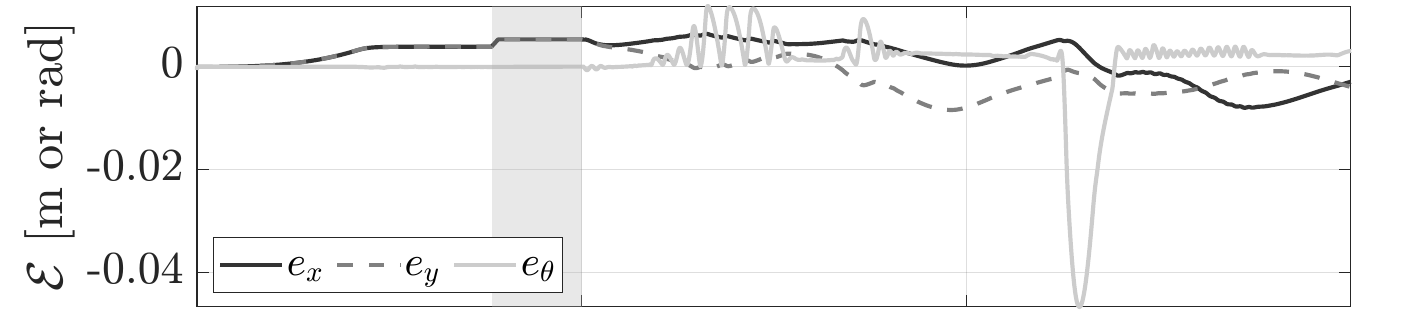} \label{2_plot_error}}
    \quad
    \subfloat[]{\includegraphics[clip, trim=0cm 0.85cm 0cm 0.8cm,width=0.48\textwidth]{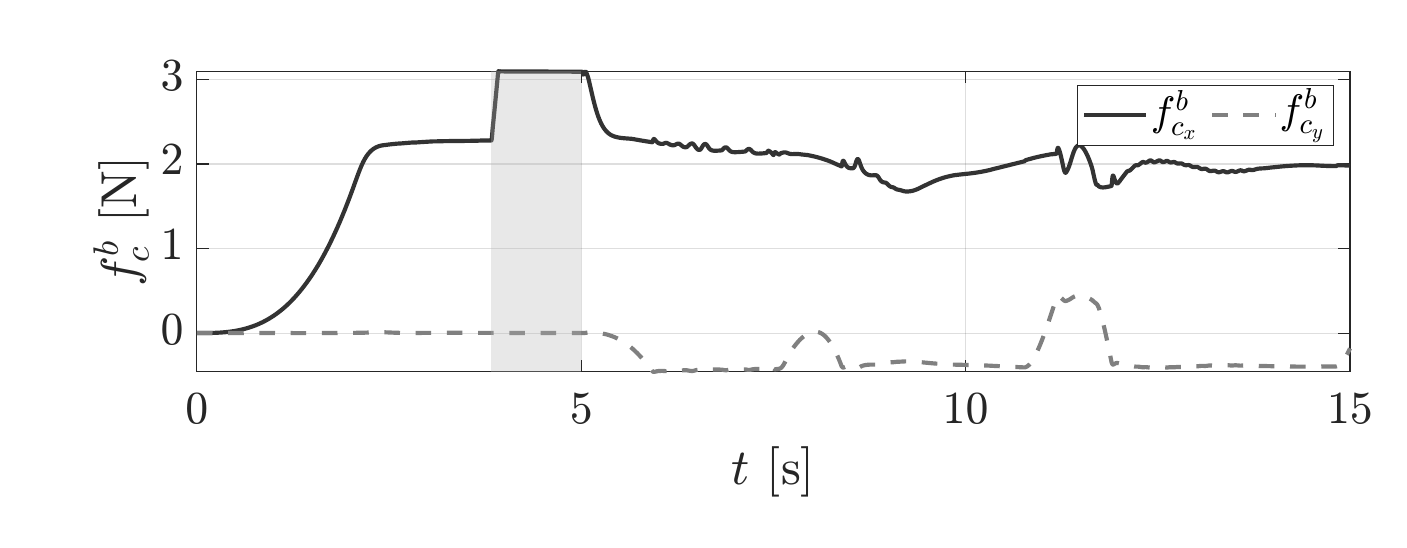} \label{3_plot_force_body}}
    \caption{Time history of the object tracking error \protect\subref{2_plot_error} and predicted forces \protect\subref{3_plot_force_body} during the \textit{passive case} simulation.}
    \label{2_plot_passive_sim}
\end{figure}
\begin{figure}[t]
    \centering
    \captionsetup[subfloat]{labelfont=scriptsize,textfont=scriptsize}
    \subfloat[]{\includegraphics[clip, trim=0cm 2.7cm 0cm 0.2cm,width=0.48\textwidth]{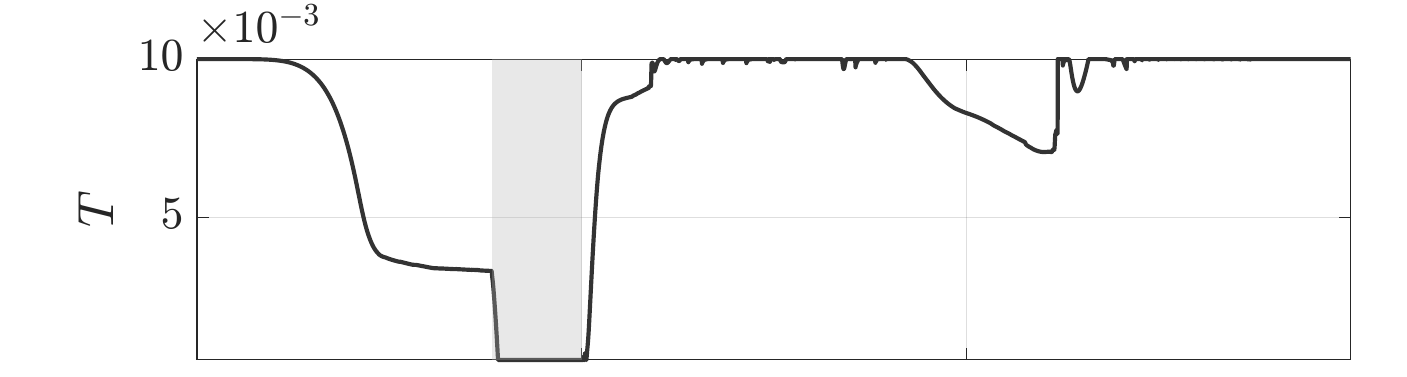} \label{2_plot_T}}
    \quad
    \subfloat[]{\includegraphics[clip, trim=0cm 0.85cm 0cm 0.8cm,width=0.48\textwidth]{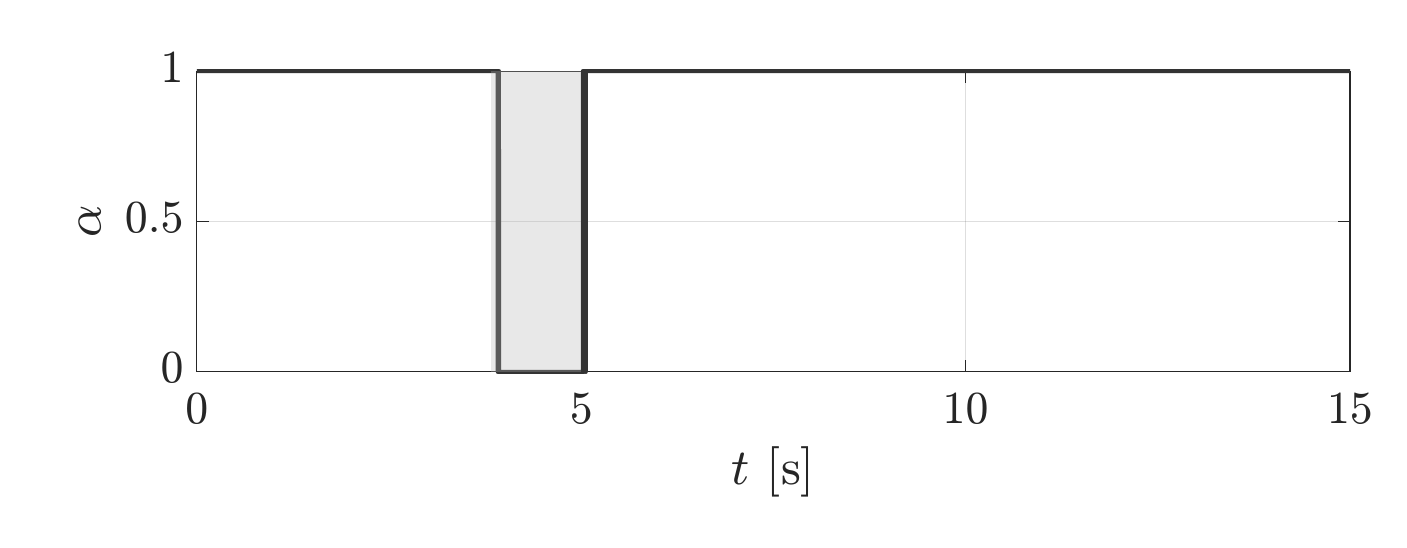} \label{2_plot_alpha_p}}
    \caption{Time history of the tank energy \protect\subref{2_plot_T} and tank activation parameter \protect\subref{2_plot_alpha_p} during the \textit{passive case} simulation.}
    \label{2_tank_state}
\end{figure}
The \textit{passive case} simulation considers an eight-shaped reference trajectory for the object, with an obstacle spawning at $t=3$~s in front of the object and disappearing at $t=5$~s. 
%
%
In this simulation, the interaction between the object and the obstacle happens between the time instants $t=3.83$~s and $t=5$~s. 
We report the results up to $t=15$~s to analyze  the system behavior during and after the occurrence of the unexpected interaction; the performance along the whole trajectory is shown in the video accompanying this paper.
Results are shown in Figs.~\ref{2_plot_passive_sim} and~\ref{2_tank_state}.
In all the plots, the time interval during which the external interaction occurs is highlighted with a shaded gray area.
The obstacle, which hinders the motion, causes an increase of the tracking error (Fig.~\ref{2_plot_error}), which consequently generates an increase of the pushing force (Fig.~\ref{3_plot_force_body}). 
This causes a reduction of the tank energy (Fig.~\ref{2_plot_T}) starting at time $t=3.83$~s when the collision happens, until the lower bound is reached at time $t=3.91$~s. At this point, the tank deactivates changes in the manipulator velocity set‐point generated by MPC via the activation parameter $\alpha$
(Fig.~\ref{2_plot_alpha_p}), preventing the pushing force from growing indefinitely. 
At time $t=5.02$~s, once the obstacle is removed, the motion is no longer hindered and the manipulator resumes its pushing movement, the tank energy starts increasing and the velocity set‐point is reactivated. 
It is possible to observe that, in terms of absolute value, both the components of the object position tracking error do not exceed $1.5 \times 10^{-2}$~m, while the orientation error remains below $10^{-2}$~rad. 
These values demonstrate the precision of our framework in executing the non‐prehensile pushing task, despite the robot operating in a compliant mode.

\begin{figure}[t]
    \centering
    \captionsetup[subfloat]{labelfont=scriptsize,textfont=scriptsize}
    \subfloat[]{\includegraphics[clip, trim=0cm 2.7cm 0cm 0.8cm,width=0.48\textwidth]{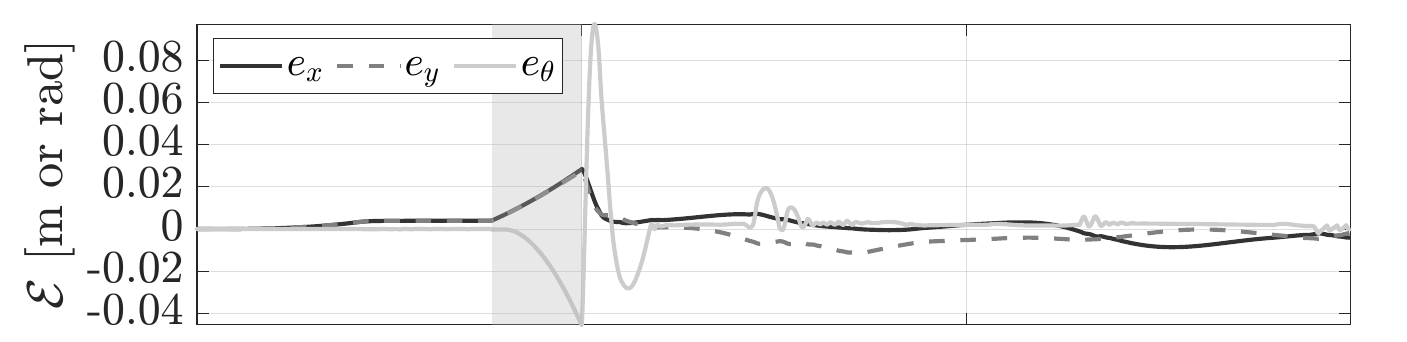} \label{3_plot_error}}
    \quad
    \subfloat[]{\includegraphics[clip, trim=0cm 0.85cm 0cm 1.1cm,width=0.48\textwidth]{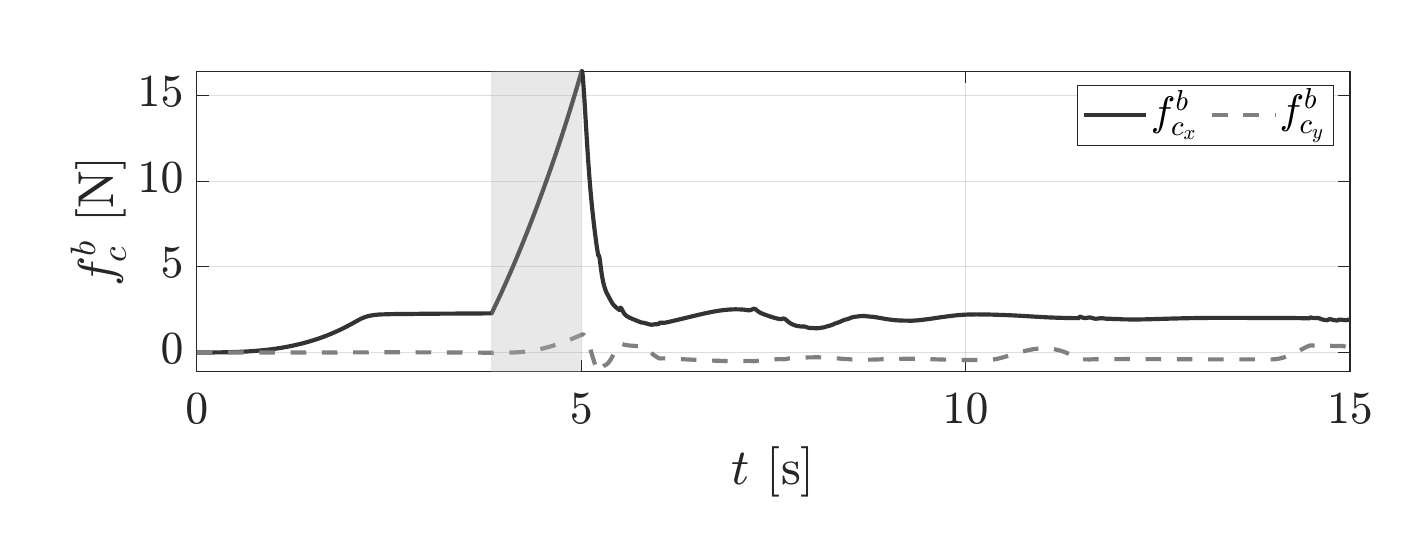} \label{3_plot_force_body_no_tank}}
    \caption{Time history of the object tracking error \protect\subref{3_plot_error} and predicted forces \protect\subref{3_plot_force_body_no_tank} during the \textit{non-passive case} simulation.}
    \label{3_plot_non_passive_sim}
\end{figure}

The \textit{non-passive case} simulation was conducted under identical conditions, but with the passivity filter removed.
Results are shown in Fig.~\ref{3_plot_non_passive_sim}. 
In this case, when the object's interaction with the obstacle occurs at time $t=3.83$~s, the tracking error begins to increase (Fig.~\ref{3_plot_error}) and the pushing force correspondingly rises in a potentially dangerous manner until the obstacle is removed (Fig.~\ref{3_plot_force_body_no_tank}). 
After the obstacle is removed, the tracking error returns to its nominal values. 

The sharp increase in force results from the controller’s attempt to reduce the growing tracking error.
In the absence of a passivity filter, no mechanism prevents the controller from continuously increasing the pushing force. 
Such behavior is clearly undesirable, as physical interactions that hinder motion are not intended to be treated as disturbances to be rejected within the proposed framework.
Therefore, integrating a passivity filter into non-prehensile pushing is essential to prevent unsafe system responses.
%


\section{Experimental Evaluation}\label{sec:experiments}
In this section, we experimentally validate the proposed control framework on real hardware, assessing whether the properties established in simulation—namely, safe passive interaction and accurate tracking under impedance control—hold under real-world conditions. 
To this end, the two complementary aspects are investigated on two robotic platforms with significantly different hardware characteristics: the passive interaction behavior of the system during physical human intervention, tested on the ABB Mobile YuMi (Section~\ref{subsec:yumi}), and the tracking performance under varying task conditions, assessed on the KUKA LBR iiwa 7 (Section~\ref{subsec:kuka}). 
%
%
%
%
The use of two platforms is motivated by their performance complementarity. 
The ABB Mobile YuMi is a platform specifically designed for human-centered environments, making it the natural choice for validating passive interaction behavior. 
However, its stringent hardware limitations restrict the achievable motions to slow velocities, preventing a thorough assessment of the controller's performance. 
The KUKA LBR iiwa 7, instead, is a sensitive, lightweight collaborative robot designed for delicate assembly work and human-robot collaboration, and is more suitable for pushing the controller to its performance limits.
The two platforms differ both in terms of maximum payload (0.5~kg for the YuMi, 7~kg for the LBR iiwa 7) and available control modalities (current interface on the YuMi, torque/position interface on the LBR iiwa 7). 
Despite these differences, the proposed controllers were implemented on both platforms with minimal modifications, achieving compliant pushing manipulation and passive interactions using only onboard sensors.

\subsection{ABB Mobile YuMi Experiments} \label{subsec:yumi}
Here we show the compliant interaction experiments performed with the ABB Mobile YuMi robot. 
The goal of these experiments is to validate the proposed framework in guaranteeing a passive behavior when an external operator physically interacts with the system during the pushing task.
\subsubsection*{A1) Experimental setup}
\begin{figure}[t]
\centering
\includegraphics[width=0.95\linewidth]{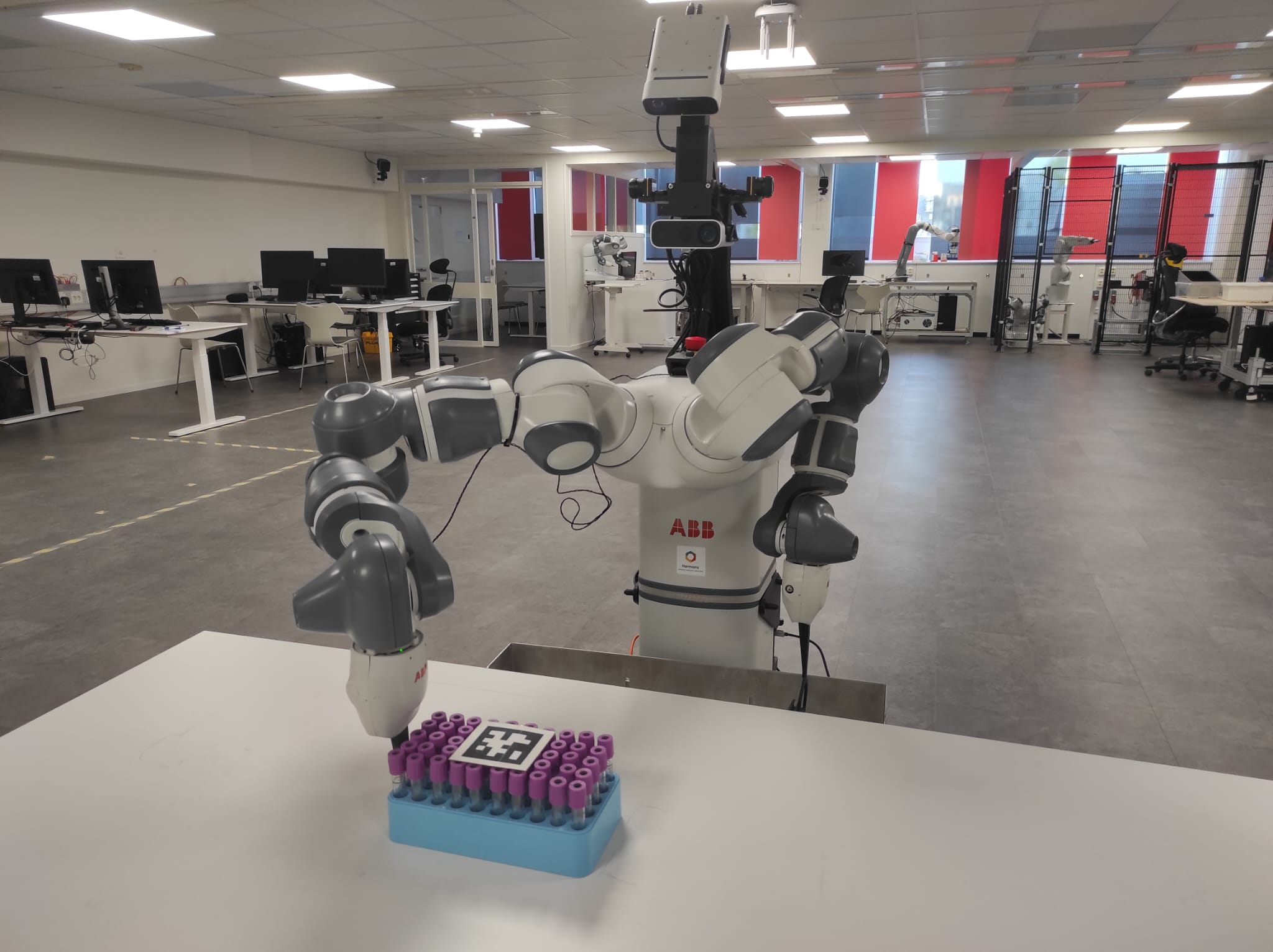}
\caption{The Mobile YuMi platform while pushing a rack of test tubes.}
\label{MobileYumi}
\end{figure}
The ABB Mobile Yumi (Fig.~\ref{MobileYumi}) is a robotic prototype
constituted by a YuMi robot (a platform composed by two 7‐DOF manipulators) mounted on top
of an omnidirectional mobile base. 
The standard interface of the YuMi does not offer the possibility to explicitly command joints' torques, nor it is equipped with torque sensors. 
To circumvent this limitation, we directly control the motors' currents to nominally set the desired torque references. 
Such interface runs at $250$~Hz. 
Additionally, our setup includes an Azure Kinect RGB-D camera operating at 15 Hz, positioned to frame the robot workspace and used to estimate object poses via AprilTags.
This solution was preferred over an external motion tracking system to demonstrate the robustness of our framework to the limited frame rate, errors, and noise associated with onboard visual perception.
During the pushing task, the orientation of the pushing tool about the vertical axis is regulated through a PD controller to keep alignment with the object.

For this experiment, we envision a scenario in which a medical professional safely interacts with a robotic assistant which pushes a rack containing test tubes for medical analysis along a desired trajectory. 
The rack, including all the tubes, has mass $m=0.474$~kg and dimensions $l_1 \times l_2 \times l_3$, with $l_1=0.21$~m, $l_2 = 0.09$~m, $l_3 = 0.10$~m. 
To perform this task, we use the robot’s right arm, equipped with a custom 3D-printed cylindrical pushing tool (with approximate height and diameter of $0.094$~m and $0.015$~m, respectively). 
Throughout the task execution, the pushing-tool pose set-points are updated at $15$~Hz to match the working frequency of the vision system. 

\begin{table*}
\centering
\caption{Parameters choice of the control framework used for experiments with the ABB Mobile Yumi.}
\label{tab:param_yumi}
\setlength{\extrarowheight}{2pt}
{
\begin{tabular}{|p{6cm}|p{6cm}|p{3.5cm}|}
\hline
\centering{\bfseries MPC}&\centering{\bfseries Robot} & \centering\arraybackslash{\bfseries Passivity filter}\\
\hline
$\boldsymbol{x}_0=[0, 0, 0, \pi, \frac{-l_1}{2}, 0, 0, 0]$ 
&
$\boldsymbol{K}_d = \operatorname{diag}\{3\mathrm{e}2,3\mathrm{e}2, 3\mathrm{e}2, 90, 90, 90\}$
&
$T_0 = 5\mathrm{e}{-3}$
\\
$\boldsymbol{w}_x = [3\mathrm{e}2, 3\mathrm{e}2, 3\mathrm{e}2, 0, 0, 0, 0, 1\mathrm{e}{-4}]$ 
&
$\boldsymbol{D}_d = \operatorname{diag}\{50,50, 50, 15, 15, 15\}$
&
$z_0 = \sqrt{2T_0}$\\
$\boldsymbol{w}_{u} = [0.1, 0.1, 0.1, 0, 0.1]$ 
&
-
&
$\bar{T} = 6\mathrm{e}{-3}$
\\
$\boldsymbol{W}_x = \operatorname{diag}\{\boldsymbol{w}_x\},\,\boldsymbol{W}_y = \operatorname{diag}\{10\boldsymbol{w}_x, \boldsymbol{w}_{u}\}$
&
-
&
$
T_{\epsilon} = 1.5\mathrm{e}{-3}
$
\\
$\nu_s = 15$ Hz, $N = 3$
&
-
&
$\boldsymbol{\varLambda} = \operatorname{diag}\{50, 50, 0, 0, 0, 0\} $
\\
\hline
\end{tabular}
}
\end{table*}

The controller parameters reported in Table~\ref{tab:param_yumi} were selected following the same rationale adopted for the simulations. However, due to the inherent difference between simulated and real-world conditions, the MPC weights used in simulation (Table~\ref{tab:param_sim}) and those employed in the real experiments (Table~\ref{tab:param_yumi}) differ substantially. 
In both cases, the object tracking error is penalized significantly more heavily than the remaining state and control variables. In simulation, this weight can be increased substantially, as the controller operates at $1$~kHz, enabling more aggressive tuning. 
In contrast, real experiments are constrained by the camera working frequency of $15$~Hz, requiring the weights to be scaled accordingly. 
Given the two-order-of-magnitude difference in control frequency, a corresponding disparity in the selected weights is expected.
The energy tank bounds also differ between the simulation setup and the real experiments. 
%
Selecting appropriate bounds is intrinsically challenging and typically requires empirical tuning. The chosen criterion is to keep the passivity filter inactive under nominal frictional forces, while ensuring rapid discharge during unexpected external interactions.
Regarding MPC initialization, the object coordinates in the world frame are biased using the measured initial position, such that the reference trajectory is translated to match the actual starting configuration. The friction coefficients between the pushing tool (plastic) and the object (plastic), and between the object and the desk surface (plastic), were both set to $0.2$. 


\subsubsection*{A2) Results} 
Two experiments are carried out with the energy tank passivity filter enabled, considering two different reference trajectories for the object: a \textit{linear trajectory} and a \textit{curvilinear trajectory}.

\begin{figure}[t]
    \centering
    \captionsetup[subfloat]{labelfont=scriptsize,textfont=scriptsize}
    \subfloat[]{\includegraphics[clip, trim=0cm 2.7cm 0cm 1.1cm,width=0.48\textwidth]{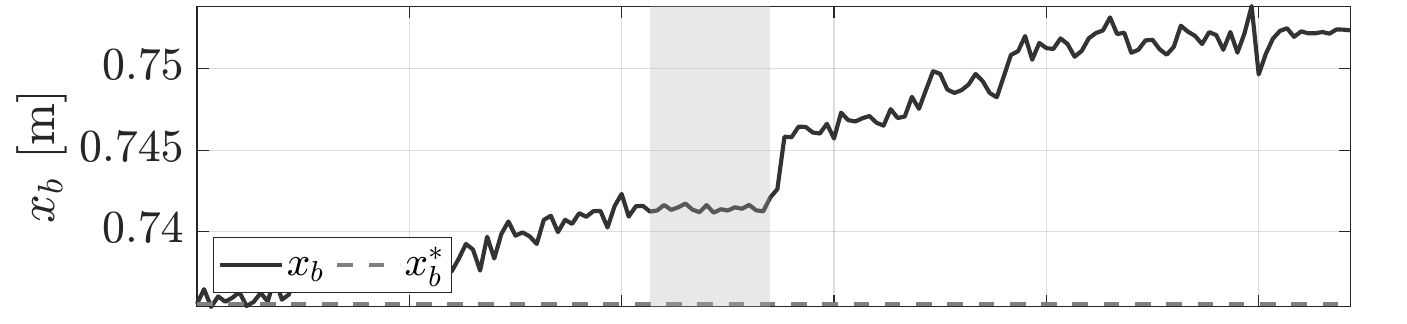} \label{4_plot_x_obj}}
    \quad
    \subfloat[]{\includegraphics[clip, trim=0cm 2.7cm 0cm 1.1cm, width=0.48\textwidth]{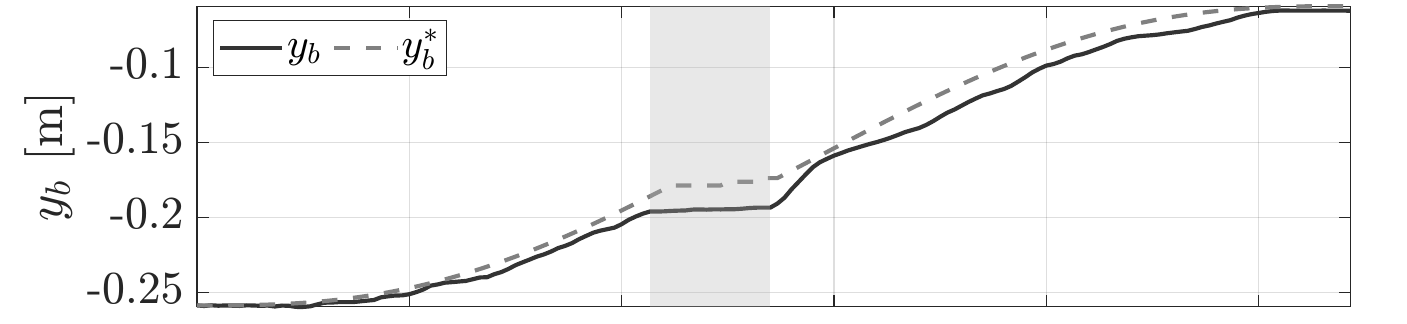} \label{4_plot_y_obj}}
    \quad
    \subfloat[]{\includegraphics[clip, trim=0cm 2.7cm 0cm 1.0cm,width=0.48\textwidth]{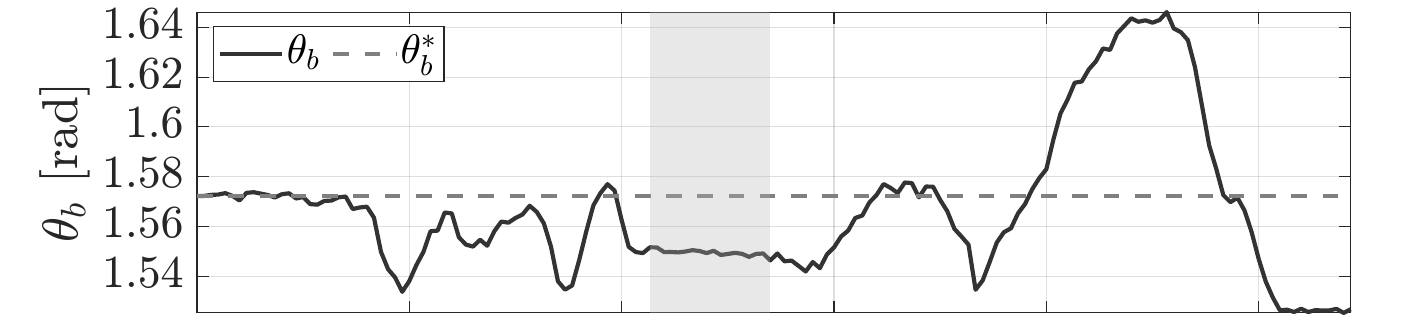} \label{4_plot_theta_obj}}
    \quad
    \subfloat[]{\includegraphics[clip, trim=0cm 0.85cm 0cm 1.1cm,width=0.48\textwidth]{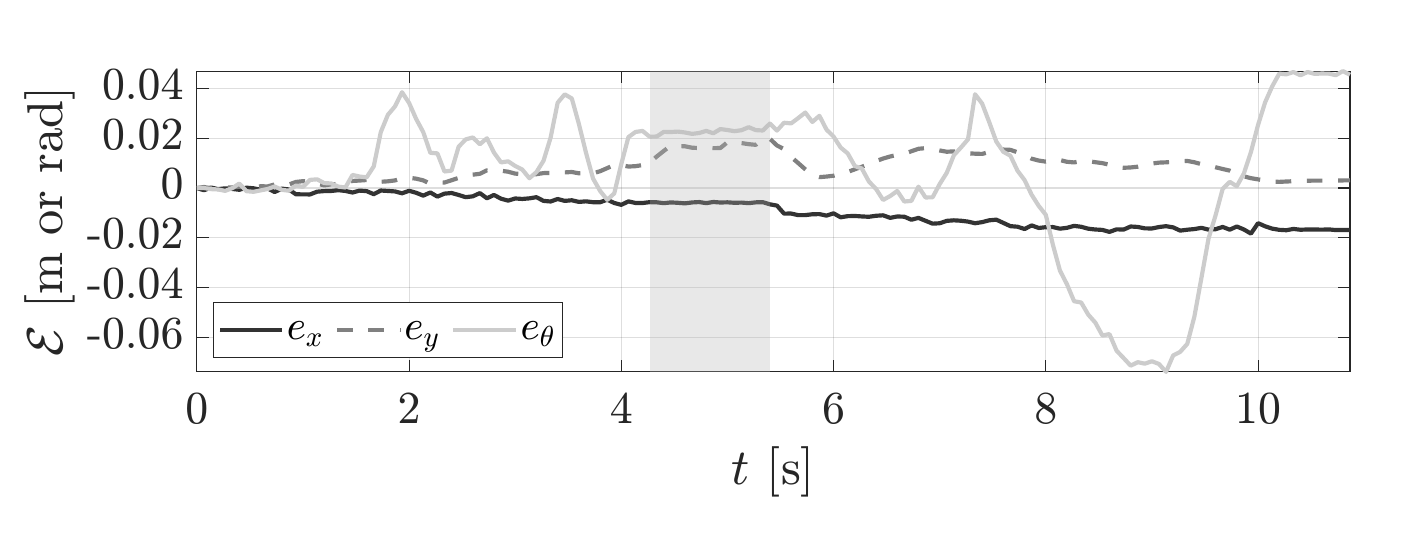} \label{4_plot_error}}
    \caption{Time history of the desired and measured object position \protect\subref{4_plot_x_obj},\protect\subref{4_plot_y_obj}, orientation \protect\subref{4_plot_theta_obj}, tracking error \protect\subref{4_plot_error} during the \textit{linear trajectory} experiment.}
    \label{4_object_state}
\end{figure}

\begin{figure}[t]
    \centering
    \captionsetup[subfloat]{labelfont=scriptsize,textfont=scriptsize}
    \subfloat[]{\includegraphics[clip, trim=0cm 2.7cm 0cm 0.2cm,width=0.48\textwidth]{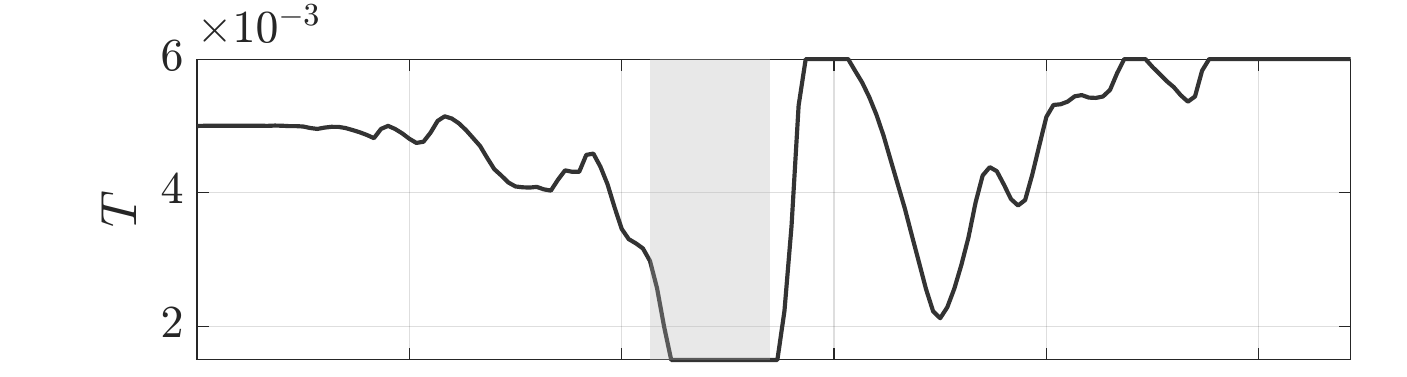} \label{4_plot_T}}
    \quad
    \subfloat[]{\includegraphics[clip, trim=0cm 0.85cm 0cm 0.8cm,width=0.48\textwidth]{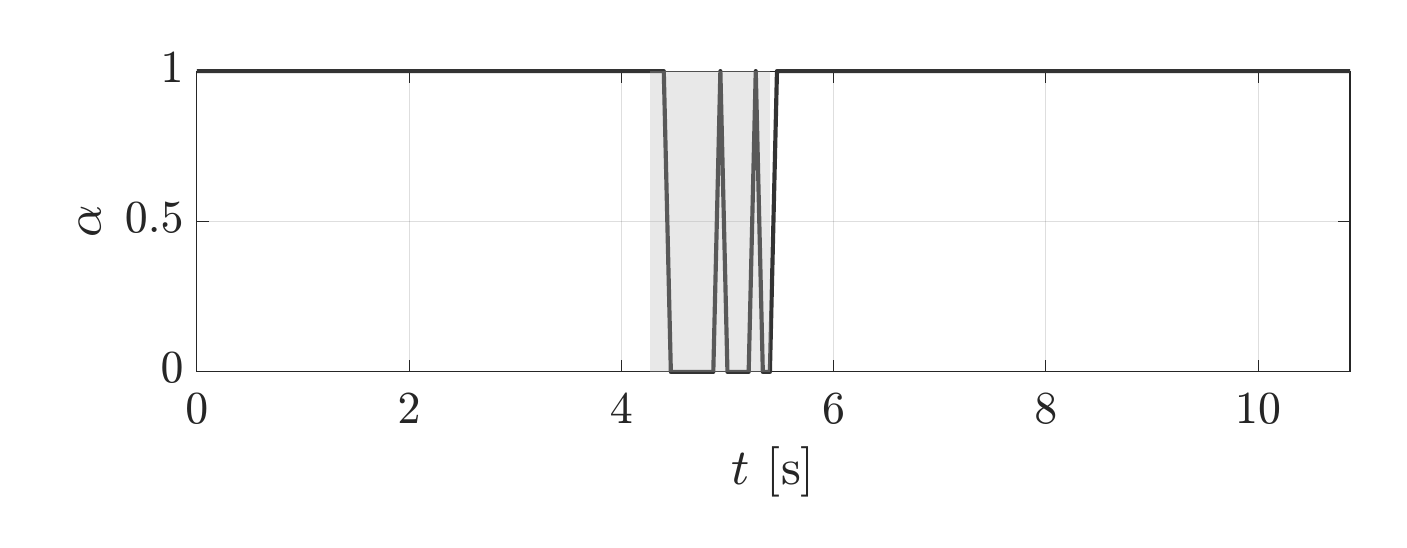} \label{4_plot_alpha_p}}
    \caption{Time history of the tank energy \protect\subref{4_plot_T} and tank activation parameter \protect\subref{4_plot_alpha_p} during the \textit{linear trajectory} experiment.}
    \label{4_tank_state}
\end{figure}

The \textit{linear trajectory} experiment results are shown in Fig.~\ref{4_object_state},~\ref{4_tank_state}.
The system maintains high-precision tracking under compliant pushing robot control with a positional error magnitude below $2 \times 10^{-3}$~m, and the orientation error below $6 \times 10^{-2}$~rad. 
During this experiment, the human hinders the object motion through physical interaction between $t=4.27$~s and $t=5.40$~s.
When the interaction begins, the tank energy decreases faster until it reaches the lower bound at $t=4.47$~s and the activation parameter disables the control action.
After the human releases the object, at $t=5.47$~s, the robot resumes motion, the energy tank recharges, and the task can be completed.

\begin{figure}[t]
    \centering
    \captionsetup[subfloat]{labelfont=scriptsize,textfont=scriptsize}
    \subfloat[]{\includegraphics[clip, trim=0cm 2.7cm 0cm 0.8cm,width=0.48\textwidth]{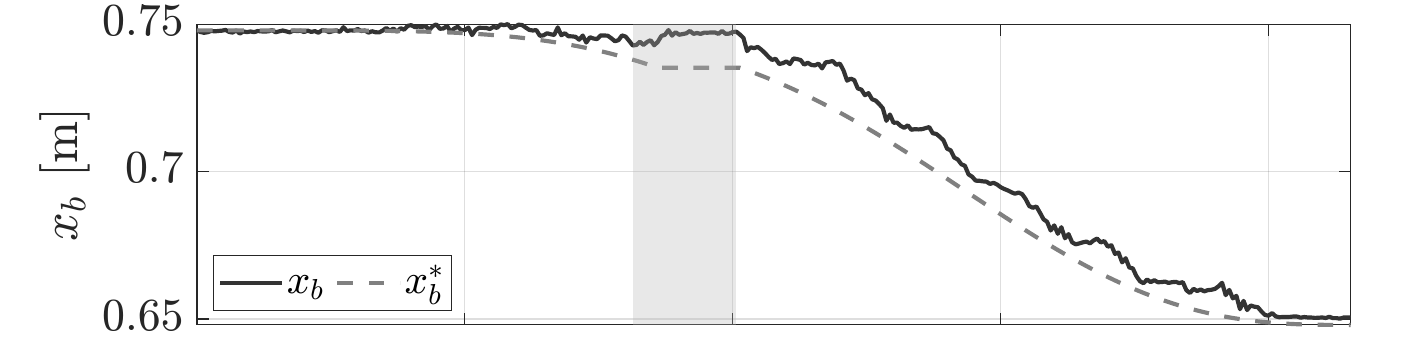} \label{5_plot_x_obj}}
    \quad
    \subfloat[]{\includegraphics[clip, trim=0cm 2.7cm 0cm 1.1cm,width=0.48\textwidth]{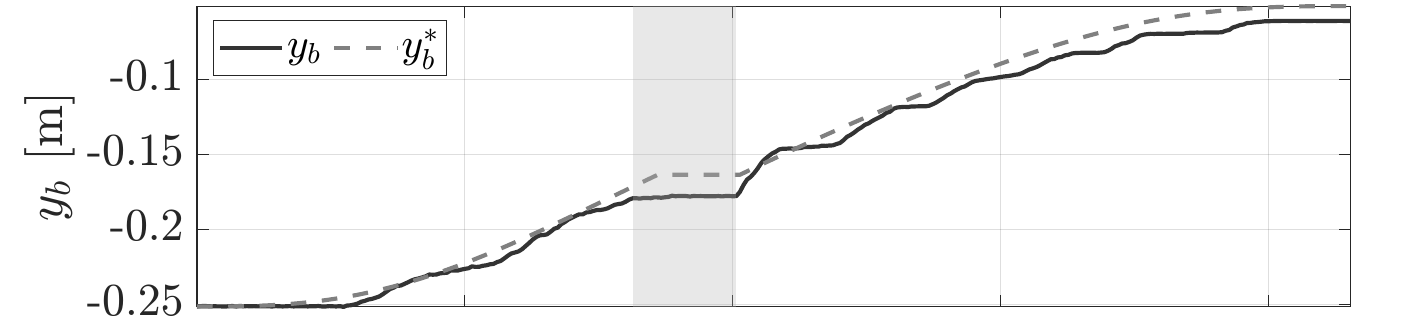} \label{5_plot_y_obj}}
    \quad
    \subfloat[]{\includegraphics[clip, trim=0cm 2.7cm 0cm 1.1cm,width=0.48\textwidth]{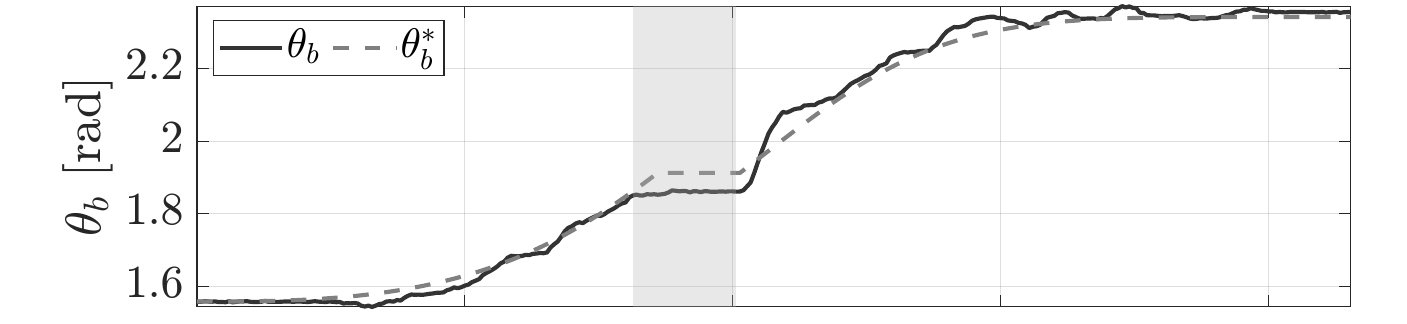} \label{5_plot_theta_obj}}
    \quad
    \subfloat[]{\includegraphics[clip, trim=0cm 0.85cm 0cm 1.1cm,width=0.48\textwidth]{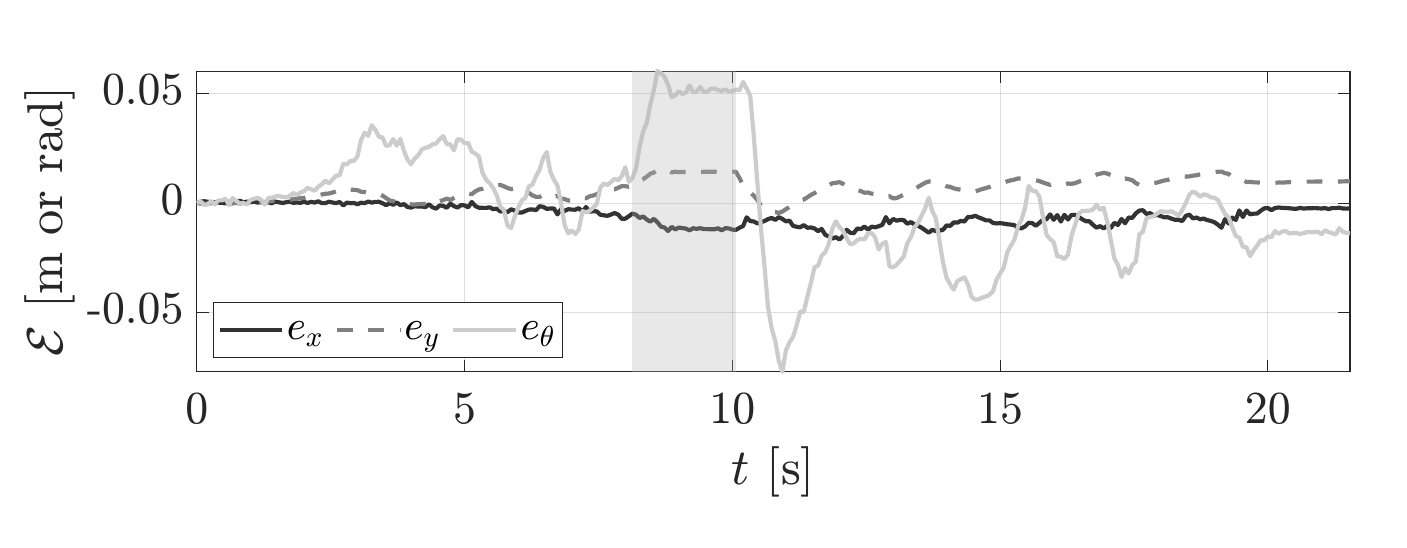} \label{5_plot_error}}
    \caption{Time history of the desired and measured object position \protect\subref{5_plot_x_obj},\protect\subref{5_plot_y_obj}, orientation \protect\subref{5_plot_theta_obj}, tracking error \protect\subref{5_plot_error} during the \textit{curvilinear trajectory} experiment.}
    \label{5_object_state}
\end{figure}

\begin{figure}[t]
    \centering
    \captionsetup[subfloat]{labelfont=scriptsize,textfont=scriptsize}
    \subfloat[]{\includegraphics[clip, trim=0cm 2.7cm 0cm 0.2cm,width=0.48\textwidth]{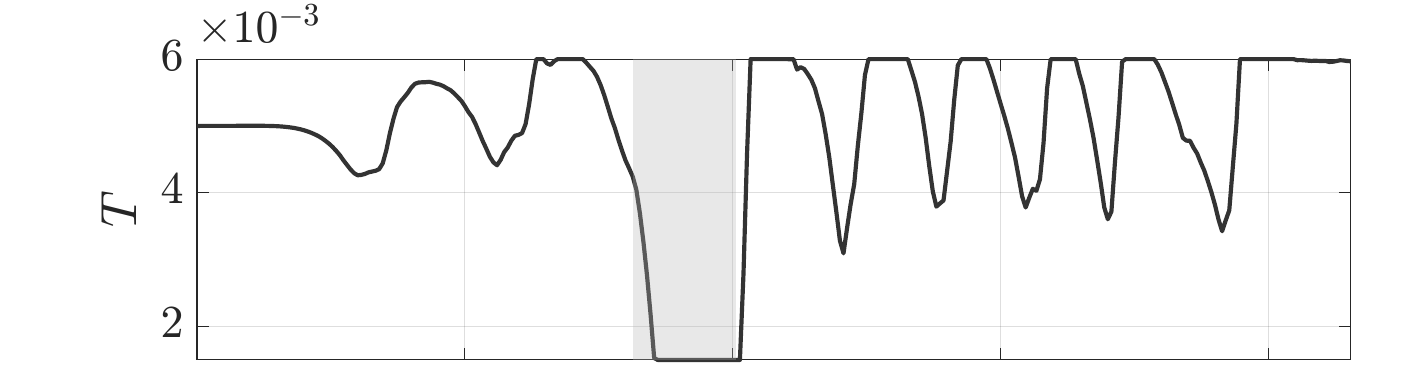} \label{5_plot_T}}
    \quad
    \subfloat[]{\includegraphics[clip, trim=0cm 0.85cm 0cm 0.8cm,width=0.48\textwidth]{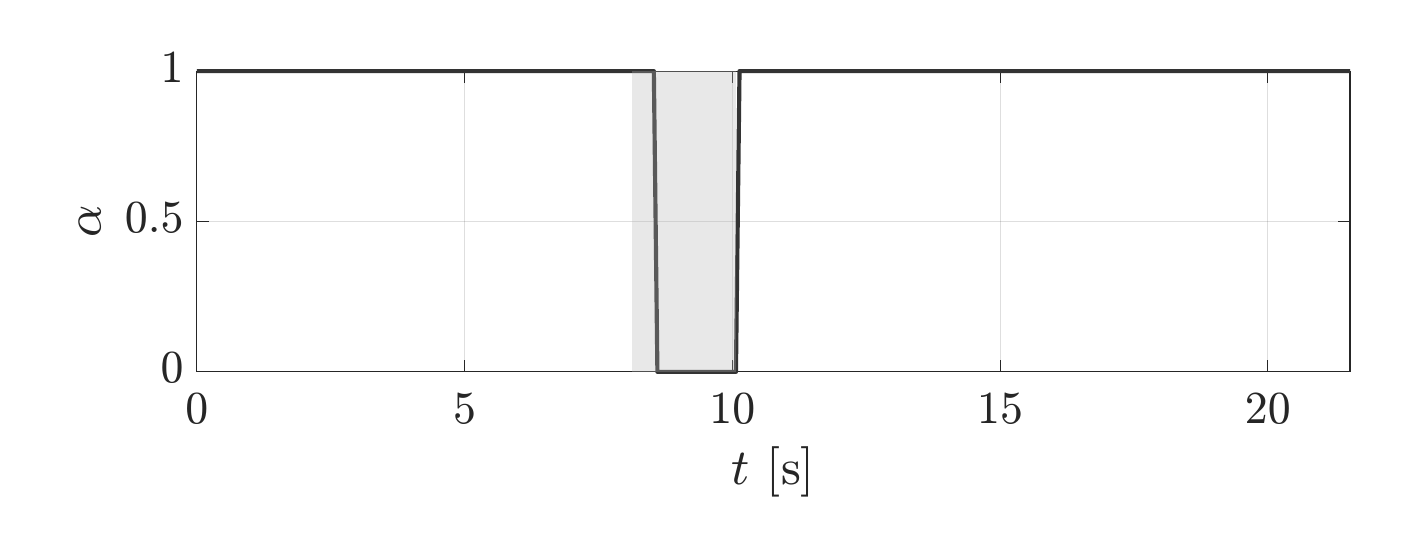} \label{5_plot_alpha_p}}
    \caption{Time history of the tank energy \protect\subref{5_plot_T} and tank activation parameter \protect\subref{5_plot_alpha_p} during the \textit{curvilinear trajectory} experiment.}
    \label{5_tank_state}
\end{figure}

The \textit{curvilinear trajectory} experiment results are reported in Fig.~\ref{5_object_state},~\ref{5_tank_state}.
The object tracking error remains on the same order of magnitude as in the previous case (positional error magnitude below $3.5 \times 10^{-3}$~m, and the orientation error below $6 \times 10^{-2}$~rad). 
In this case, the human hinders the object motion between $t=8.13$~s and $t=10.06$~s.
Likewise, the tank energy decreases until it reaches its lower bound at $t=8.60$~s and the activation parameter disables the control action. 
Once the object motion is no longer hindered, at $t=10.13$~s, the tank recharges storing dissipated energy and the robot can resume the task.

In both cases, the passive behavior with respect to the external interaction has been demonstrated.

\subsection{KUKA LBR iiwa Experiments}\label{subsec:kuka}

Here we show a quantitative experimental analysis performed to assess the tracking and the robustness capabilities of the proposed controller when pushing objects of different masses, on different surfaces, along different paths, with different velocity profiles.
Repeated experiments were performed under a variety of possible conditions to better demonstrate the performance of the presented architecture.

\subsubsection*{B1) Experimental Setup}
The KUKA LBR iiwa 7 is a 7-DOF lightweight collaborative robot designed for precise assembly tasks to be performed alongside humans. 
The Fast Robot Interface (FRI) allows commanding the robot by sending joint torque/position commands rates up to $1$-KHz. 
The control framework is developed exploiting the ROS 2 LBR-Stack\footnote{\url{https://github.com/lbr-stack/lbr_fri_ros2_stack}} wrapping the FRI~\cite{Huber2024}. 
We use the built-in \emph{Cartesian Impedance Control Mode}, which allows the user to easily set the robot stiffness and damping. 
Our setup comprises an in-hand Intel RealSense depth camera D435i, operating at $30$~Hz, to track the object's pose via AprilTags.
The robot is equipped with a custom pushing tool providing a camera support and an aluminium stick with a spherical-shaped end of radius $r = 0.0125$~m. 
The MPC problem is solved at $25$~Hz, matching the update rate of the camera. Between consecutive MPC solutions, the generated set-points are linearly interpolated at $200$~Hz to ensure smooth robot motion. The FRI interface commands the robot at the same rate of $200$~Hz.
%
As in~\cite{MouraICRA2022}, the MPC comprises $N=25$ samples in the prediction horizon.
We solve the OCP in $4.10$~ms (mean value) $\pm 1.80$~ms (std), which is  suitable for our application. 
The experiments have been executed running the controller on a machine with x86-64 architecture, powered by an AMD Ryzen 5 5600H processor with Radeon Graphics running at 1300 MHz, and operating on Ubuntu 20.04.6 LTS.
In this session of experiments, we push a rack containing test tubes of the same dimensions of those used previously. Also in this case,  the orientation of the pushing tool about the vertical axis is regulated through a PD controller to keep alignment with the object.
%
\begin{table*}
\centering
\caption{Parameters choice of the control framework used for experiments with the KUKA LBR iiwa 7.}
\label{tab:param_kuka}
\setlength{\extrarowheight}{2pt}
{
\begin{tabular}{|p{6cm}|p{6cm}|}
\hline
\centering{\bfseries MPC}&\centering\arraybackslash{\bfseries Robot}\\
\hline
$\boldsymbol{x}_0=[0, 0, 0, \pi, \frac{-l_1}{2}, 0, 0, 0]$ 
&
$\boldsymbol{K}_d = \operatorname{diag}\{3\mathrm{e}2,3\mathrm{e}2, 3\mathrm{e}2, 3\mathrm{e}2, 3\mathrm{e}2, 3\mathrm{e}2\}$
\\
$\boldsymbol{w}_x = [1\mathrm{e}2, 1\mathrm{e}2, 1\mathrm{e}2, 0, 0, 0, 1\mathrm{e}{-4}, 1\mathrm{e}{-4}]$ 
&
$\boldsymbol{D}_d = \operatorname{diag}\{0.7,0.7, 0.7, 0.7, 0.7, 0.7\}$
\\
$\boldsymbol{w}_{u} = [0.1, 0.1, 0.5, 0.5, 1\mathrm{e}{-2}]$ 
&
-
\\
$\boldsymbol{W}_x = \operatorname{diag}\{\boldsymbol{w}_x\},\,\boldsymbol{W}_y = \operatorname{diag}\{10\boldsymbol{w}_x, \boldsymbol{w}_{u}\}$
&
-
\\
$\nu_s = 25$ Hz, $N = 25$
&
-
\\
\hline
\end{tabular}
}
\end{table*}


The controller parameters are reported in Table~\ref{tab:param_kuka}. 
These values were obtained using the same tuning procedure adopted in the previous experiments, namely iterative adjustments aimed at balancing performance and stability.
Since these trials focus exclusively on performance evaluation and do not involve physical interaction scenarios, passivity filter parameters are not considered. 
In this setup, the damping parameters in Table~\ref{tab:param_kuka} are normalized; the friction coefficient between the end-effector (plastic) and the object (plastic) is  $0.2$, whereas that between the object and the aluminum table is $0.35$. 
\begin{figure*} 
    \centering
    \includegraphics[width=0.9\linewidth]{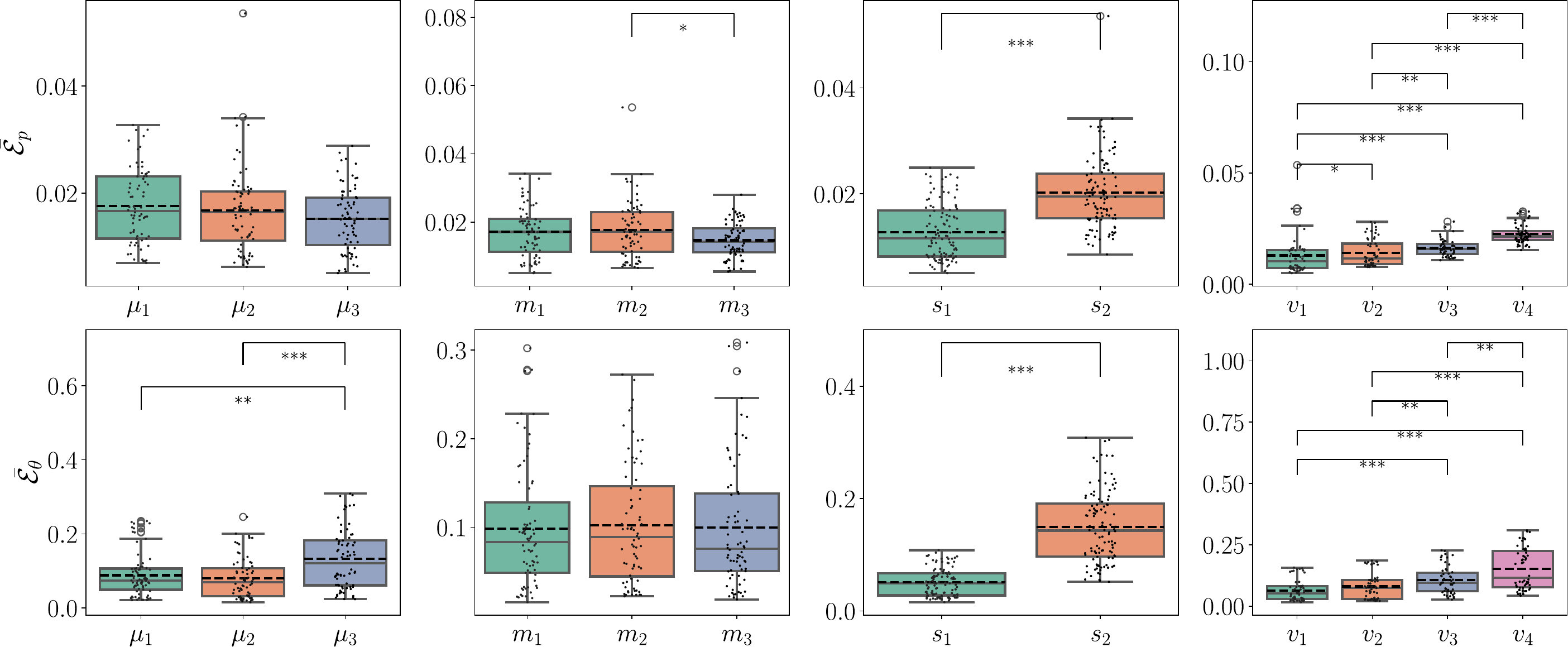}
    \caption{Box plots showing the aggregated results of the experimental campaign conducted to assess the robustness of the method. $\bar{\mathcal{E}}_p$ and $\bar{\mathcal{E}}_\theta$ denote positional and rotational RMSE. Factors are: object-table friction $\mu$, object mass $m$, desired path $s$, trajectory mean velocity $v$. 
    One (*), two (**), three (***) asterisks denote significance levels with p-value $p < 0.05,\, 0.01,\, 0.001$, respectively. 
    From left to right, it can be noted that
    object positional error is largely unaffected by friction, while rotational error increases on the high-friction surface. 
    Object mass has minimal influence, except for a slight positional improvement at higher mass. 
    Curvilinear paths degrade tracking accuracy and, finally, higher desired velocities lead to increased errors.}
    \label{fig:real_perf_statistics} 
\end{figure*}
\subsubsection*{B2) Results}
The experimental campaign was conducted considering the following factors (independent variables) and their respective levels:
\begin{itemize}
    \item[$\mu$:] Linear object-table friction coefficient: three levels corresponding to aluminum ($\mu_1 = 0.35$), nylon ($\mu_2 = 0.28$) and steel ($\mu_3 = 0.33$) tables.
    \item[$m$:] Object mass: three levels denoting the rack condition \textit{empty} ($m_1=0.1649$~kg), \textit{half-full} ($m_2 = 0.2843$~kg), and \textit{full} ($m_3 = 0.4097$~kg) of test tubes, respectively. 
    \item[$s$:] Object desired path: two levels corresponding to linear ($s_1$) and curvilinear ($s_2$) paths.
    \item[$v$:] Desired trajectory mean velocity: four levels chosen as $v_1 = 0.0313$~m/s, $v_2 = 0.0391$~m/s, $v_3 = 0.0521$~m/s, $v_4 = 0.0782$~m/s. 
\end{itemize}
The values of $v$ were selected by performing experiments at increasing speeds up to $0.078$~m/s, pushing the controller to its performance and real-world applicability limits.
Given the envisioned task/scenario, we considered the performance as acceptable if the object positional and rotational root mean square errors (RMSE) are below $0.015$~m and $0.2$~rad, respectively, in the nominal working conditions. 
In experiments conducted at increasing mean velocities, we observed that at $0.078$~m/s the adopted performance bounds are no longer satisfied.

Three experiments were run for each combination of the factors' levels leading to a total of $216$ tests.
In all the considered cases, the controller parameters are kept at their nominal level, to demonstrate the robustness of the method with respect to environmental uncertainty.
We use the collected data to analyze two object trajectory tracking performance metrics of the proposed controller, i.e., the root mean square positional ($\bar{\mathcal{E}}_p$) and rotational ($\bar{\mathcal{E}}_\theta$) errors. 
Box plots showing the aggregated data are shown in Fig.~\ref{fig:real_perf_statistics}, where the mean (dashed line) and the median (continuous line) are also reported. 
The Mann–Whitney U test was used to test whether the changes in the metrics, due to changes in the factors, are statistically significant. 
The test was chosen due to non-Gaussian distribution of the data according to the D’Agostino and Pearson’s~\cite{dagostino1, dagostino2} test that combines skew and kurtosis to produce an omnibus test of normality.
One, two, three asterisks denote significance levels corresponding to the p-value $p < 0.05,\, 0.01, \, 0.001$, respectively.

As can be noted, changes in the friction coefficient $\mu$ do not cause a statistically significant change in the object positioning error $\bar{\mathcal{E}}_p$, while for the object rotational error $\bar{\mathcal{E}}_\theta$ we observe a statistically significant change between $\mu_2$ and $\mu_3$ ($p < 0.001$) and between $\mu_1$ and $\mu_3$ ($p < 0.01$). 
This is expected and reflects greater difficulties of our controller to precisely steer the object to the desired value in case of steel table, which exhibits greater rotational friction compared to the others. 
Nevertheless, the RMSE is contained and satisfies the application requirements.

Changes in the mass of the object do not always lead to statistically significant changes in the considered metrics, showing the robustness of our method with respect to changes in the envisioned working conditions (rack empty, half-full, and full).
A statistically significant change is observed between $m_2$ and $m_3$ for the metric $\bar{\mathcal{E}}_p$ ($p<0.05$), with greater mass leading to less positional error.

In contrast, changing the desired path for the object leads to a statistically significant difference in the tracking performance along both metrics ($p < 0.001$), with the curvilinear path exhibiting higher errors. 
This denotes a strong dependence of the considered RMSEs on the choice of the pushing path: in our experience, choosing a feasible path is essential to mitigate the error and maintain a high-level task performance.

Ultimately, with a few exceptions, the tracking performance is in different measures significantly affected by the object desired mean velocity, with higher velocities leading in general to more degraded performance. 
This was also intuitively expected. 
However, in view of the considered collaborative application scenario, we expect our method to operate in slow velocity conditions.

In summary, with a general mean positional error of $1.65 \times 10^{-2} $~m and mean rotational error of $0.10$~rad in all the perturbed cases, our method exhibits performance comparable to recent pushing approaches (e.g.,~\cite{FedericoRAL2025}) where RMS positional errors of $5.2 \times 10^{-3}$~m (mean x-axis and y-axis) and rotational error
of $0.21$~rad are reported, despite the re-created uncertainty of the environment.
The slightly higher positional error achieved in our experimental setup is mainly attributable to the use of the impedance control mode, whose trajectory tracking capability is more sensitive to disturbances (e.g. static and non-homogeneous friction). 

\section{Discussion and Future Directions}
\label{discussion}
Our control framework combines two key components: an extended compliant manipulation model that enables model-predictive non-prehensile pushing with an impedance-controlled robot, and an energy tank passivity filter that guarantees system passivity. 
Several design choices and future developments are discussed here.
We discuss possible modifications of the object interaction model, the inclusion of force measurements, and tuning aspects for both the controller and the passivity filter.

\subsection{Dynamic interaction model for compliant pushing}
The proposed MPC leverages a compliant pushing model to indirectly realize a predicted force, via a spring position/velocity set-point and contact point adaptation. 
Given the envisioned quasi-static conditions, featuring low velocities and negligible accelerations, the limit surface maps this force into a resulting object velocity.  
In more dynamic conditions, the use of a spring-damper model for the force could, instead, yield an acceleration set-point (influenced by the force derivative and the second time derivative of $\varphi_b$), which would become the control input in~\eqref{control_input}, requiring an extension of the state in~\eqref{state} that should include velocity set-point and $\varphi_b$ time derivative.
This may lead to a second-order dynamic pusher–slider model suitable for pushing tasks involving non-negligible accelerations, albeit at the cost of increased MPC complexity and a consequently more delicate controller tuning procedure.
We deliberately omitted this aspect in our controller design, adopting a trade-off that is central to our MPC framework: the model should not be more complex than required by the task.
In our quasi-static scenario, the use of the simpler model is justified and is shown to provide satisfactorily robust performance under parameter variability.
As future work, it would be interesting to address the problem of faster, more dynamic pushing tasks, also considering safe physical interaction in such regimes, as well as generalizing or quantifying robustness to object shape variation~\cite{FedericoRAL2025}.

\subsection{Force feedback}
In both simulations and experiments conducted using our MPC framework, force updates rely on predictions, which proved adequate for the considered task.
Since force derivatives act as control inputs while force itself is included in the system state, it could alternatively be updated using measured force values, provided that a force sensor is available.
This is not trivial, as the quasi-static pusher–slider model uses the mapping in~\eqref{lim_sur}, which does not uniquely determine the force intensity for a given velocity. 
Improved accuracy could be obtained via online identification of the limit surface, at the cost of a time-varying model which may potentially impact the MPC convergence.

%
In our current design, the predicted force is also employed in the energy tank passivity filter in~\eqref{pass_fil_compact}.
Replacing it with measured force feedback would require separating the robot-applied component from external interactions (see Section~\ref{passivity}). In some cases, appropriately placed force or tactile sensors at the object interface may provide this information in isolation.
With suitable modifications, our framework could therefore be extended to incorporate measured forces, potentially improving both control performance and safety through real-time force feedback.


\section{Conclusion}
We proposed an MPC framework for non-prehensile pushing that ensures passive and compliant behavior of the robot with respect to external interactions. Our extension of the state-of-the-art optimal control formulation in~\cite{MouraICRA2022} enabled an impedance-controlled robot to achieve a desired object motion by jointly optimizing the pushing force and contact point evolution through the robot's position and velocity set-point.
An energy tank passivity filter was introduced to prevent loss of passivity when a human hinders the manipulative motion, a condition that would otherwise cause the controller to increase the pushing force.
The proposed framework enables the safe execution of non-prehensile pushing tasks in human-centered environments, by explicitly handling  external interactions, whether intentional or accidental.
Results obtained from physics-based simulations and real-world experiments support these claims and further demonstrate the robustness of the controller under varying operating conditions. 
The significance of these findings has been discussed, and potential directions for future extensions have been outlined.
\section*{Acknowledgments}
The authors wish to express their sincere gratitude to Prof. Luigi Villani for his thorough review and valuable remarks, which played an important role in improving the clarity and quality of this manuscript.
\bibliographystyle{ieeetr}
\bibliography{bibliography}

\begin{IEEEbiography}[{\includegraphics[width=1in,height=1.25in,clip,keepaspectratio]{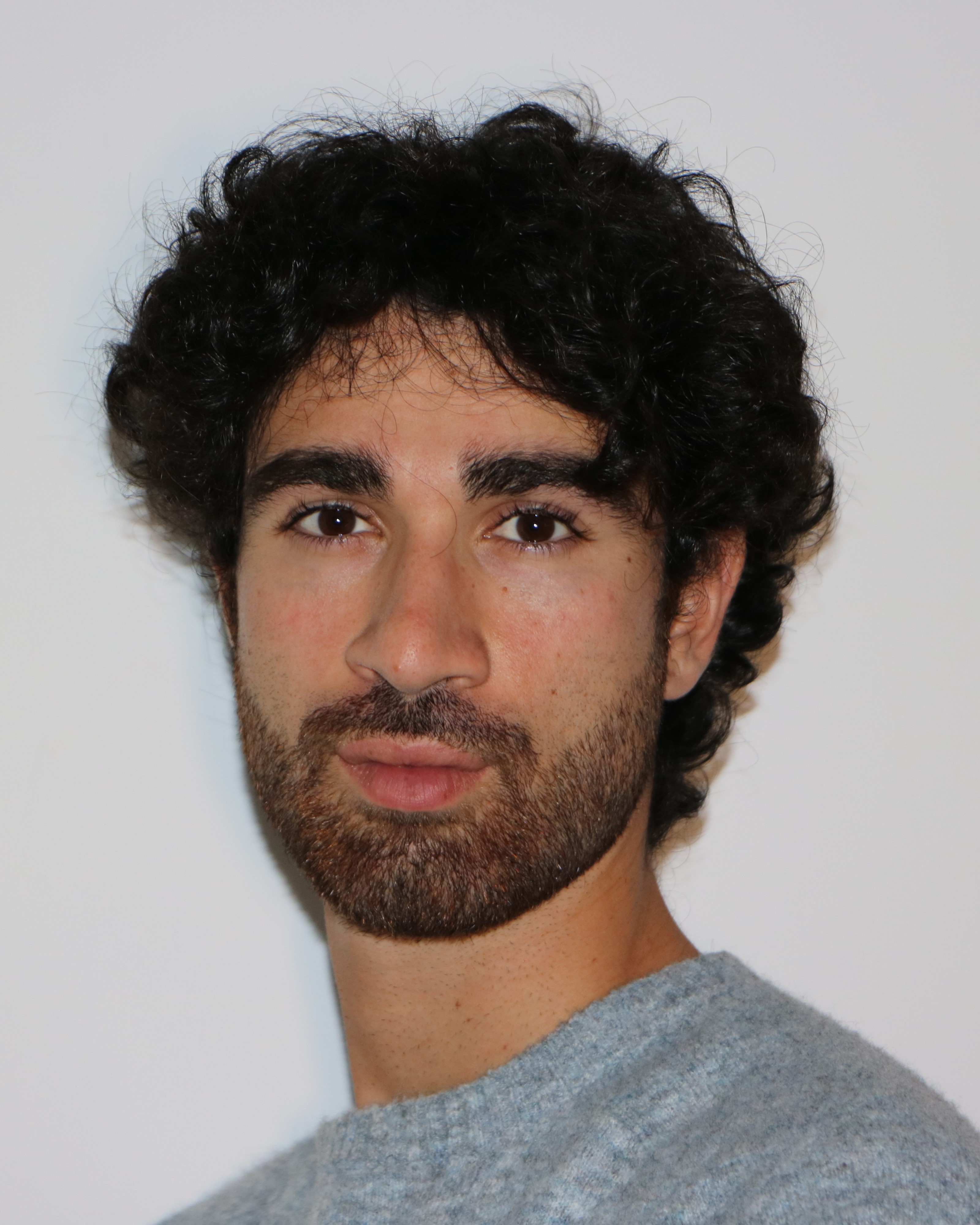}}]{Francesco Cufino}
	is a PhD student with the department of Electrical Engineering and Information Technology of the University of Naples Federico II, Naples, Italy.  From the same institution, he received the bachelor's degree in Automation Engineering in 2020, and the master's degree in Automation Engineering and Robotics in 2023. His research interests are focused on control strategies for robotic manipulation.
\end{IEEEbiography}

\begin{IEEEbiography}[{\includegraphics[width=1in,height=1.25in,clip,keepaspectratio]{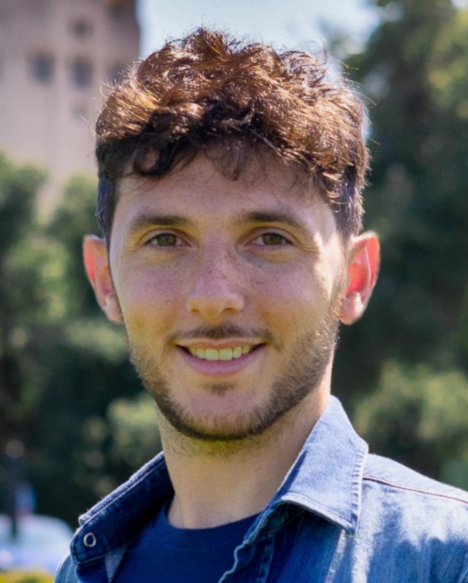}}]{Mario Selvaggio}
	(S’16, M’20) is a Tenure Track Researcher with the Department of Electrical Engineering and Information Technology, University of Naples Federico II, Naples, Italy. He received the bachelor’s and master’s degrees in mechanical engineering in 2013 and 2015, and the Ph.D. in information technology and electrical engineering in 2020, all from the same institution. From 2020 to 2023 he was a Postdoctoral Researcher. He serves as Associate Editor for the IEEE Transactions on Robotics, the IEEE/ASME Transactions on Mechatronics, and the IEEE Robotics and Automation Letters. He has authored more than 50 peer-reviewed journal and conference papers and book chapters.
\end{IEEEbiography}

\begin{IEEEbiography}[{\includegraphics[width=1in,height=1.25in,clip,keepaspectratio]{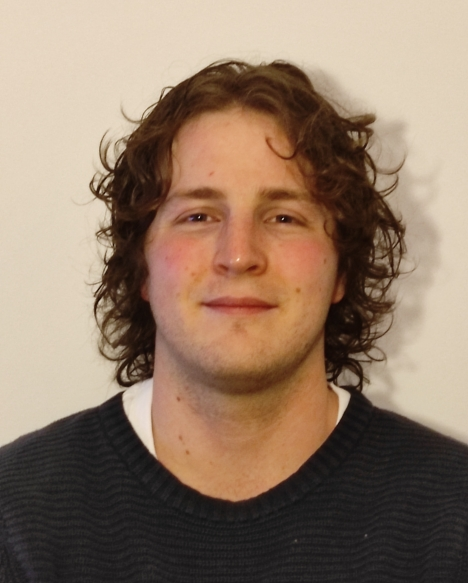}}]{Fabio Amadio}
is a Postdoctoral Researcher at the Inria Centre at Université de Lorraine, Nancy, France. He received the M.Sc. degree in automatic control and robotics from the Technical University of Catalonia (UPC), Barcelona, Spain, in 2018, and the M.Sc. degree in automation engineering and the Ph.D. degree in information engineering from the University of Padua, Padua, Italy, in 2018 and 2022, respectively. He was a Research Fellow with Leonardo Labs and the Istituto Italiano di Tecnologia, Genoa, Italy (2021–2023) and a Research Scientist with ABB Corporate Research, Västerås, Sweden (2023–2024). His research interests are focused on learning-based methods for robot control.
\end{IEEEbiography}

\begin{IEEEbiography}[{\includegraphics[width=1in,height=1.25in,clip,keepaspectratio]{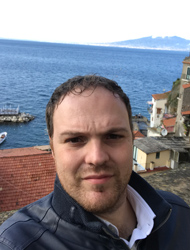}}]{Fabio Ruggiero} (S’07–M’10–SM’19)  is Associate Professor of Automatic Control and Robotics in the Department of Electrical Engineering and Information Technology at University of Naples Federico II, where he is responsible for the DynLeg (Dynamic manipulation and Legged robotics) research area. He received the M.Sc. degree in Automation Engineering in 2007 and he got the Ph.D. degree in 2010. His research interests are focused on model-based control design of robotic systems. In particular, his studies are specialized on control strategies for dexterous, dual-hand and nonprehensile robotic manipulation, aerial robots, aerial manipulators, and legged robots. He is Chair of the IEEE Italy RAS Chapter. He has co-authored more than 110 among journal papers, book chapters, and conference papers.
\end{IEEEbiography}

\end{document}